%% file: arxiv_paper.tex
\documentclass{article}
    \usepackage{arxiv}
    \pdfoutput=1

        \usepackage{algorithm, algpseudocode}
        \usepackage{amsmath, amsfonts}
        \usepackage{graphicx, wrapfig, caption, subcaption}
        \usepackage{bm}
        \usepackage[hidelinks]{hyperref}
        \usepackage[utf8]{inputenc}
        \usepackage{listings}
        \usepackage{booktabs}
        \usepackage[dvipsnames]{xcolor}
    
        \newcommand{\beps}{\boldsymbol{\epsilon}}
        \newcommand{\B}{\mathbf{B}}
        \newcommand{\M}{\mathbf{M}}
        \newcommand{\mybar}{\mid}
        \newcommand{\obs}{\bx^{o}}
        
        \newcommand{\prior}{\pi(\boldsymbol{\theta})}
        \newcommand{\bt}{\boldsymbol{\theta}}
        \newcommand{\bb}{\boldsymbol{\beta}}
        \newcommand{\bthat}{\boldsymbol{\hat{\theta}}}
        \newcommand{\bx}{\mathbf{x}}

        \newcommand{\eps}{\epsilon}
        
        \newcommand{\post}{p(\bt \mybar \obs)}
        \newcommand{\sumstats}{\mathbf{\Psi}}

        \newcommand{\ud}[1]{\mathop{d#1}}
        \newcommand{\Xtheta}{\mathcal{X}_{\theta}}
        \newcommand{\xtheta}{\bx_{\theta}}
        \newcommand{\Xm}{\mathcal{X}_m}
        \newcommand{\xm}{\bx_{m}}
        \newcommand{\ntheta}{n_{\theta}}
        \newcommand{\nm}{n_m}
        \newcommand{\hbeta}{h_{\bb}}
        \newcommand{\betahat}{\boldsymbol{\hat{\beta}}}

    \usepackage{graphicx}
        
        \usepackage[round,authoryear,sort]{natbib}
        \setcitestyle{aysep={}} 
        \newcommand{\textcite}{\citet}

        \newenvironment{traizq}[3]
        {
            \gdef\mycaption{#1}
            \gdef\mylabel{#2}
            \gdef\mylofcaption{#3}
    
            \begin{table}[tb]
                \vskip 3mm
                    \begin{center}
                        \begin{small}
                            \begin{sc}
                                \begin{tabular}{lccccccr}
                                    \hline
                                
        }
        {
                                \hline
                            \end{tabular}
                        \end{sc}
                    \end{small}
                \end{center}
                \caption[\mylofcaption]{\mycaption}
                \label{tab:\mylabel}
                \vskip -3mm
            \end{table}
        }
        \newenvironment{traizq*}[3]
        {
            \gdef\mycaption{#1}
            \gdef\mylabel{#2}
            \gdef\mylofcaption{#3}
    
            \begin{table*}[tb!]
                \vskip 3mm
                    \begin{center}
                        \begin{small}
                            \begin{sc}
                                \begin{tabular}{lccccccr}
                                    \hline
                                
        }
        {
                                \hline
                            \end{tabular}
                        \end{sc}
                    \end{small}
                \end{center}
                \caption[\mylofcaption]{\mycaption}
                \label{tab:\mylabel}
                \vskip -3mm
            \end{table*}
        }

\begin{document}

    \twocolumn[
      \aistatstitle{Dynamic Likelihood-free Inference via Ratio Estimation
        (DIRE)}
    \aistatsauthor{ Traiko Dinev \And Michael U. Gutmann}
    \aistatsaddress{School of Informatics \\[-1ex] University of Edinburgh\\[-1ex]  \href{mailto:traiko.dinev@ed.ac.uk}{traiko.dinev@ed.ac.uk} \And School of Informatics\\[-1ex] University of Edinburgh\\[-1ex] \href{mailto:michael.gutmann@ed.ac.uk}{michael.gutmann@ed.ac.uk}}]

    \input{main_body}

    \clearpage
    \clearpage
    
    \setcitestyle{numbers}
    \bibliography{nn,abc}
    
    \clearpage
    \begin{appendix}
      \onecolumn
      \thispagestyle{empty}
      \begin{center}
        {\Large \bfseries Supplementary Material}\\
        \noindent\rule{\textwidth}{1.5pt}
      \end{center}
    \input{supplementary}
    \end{appendix}
\end{document}

%% file: main_body.tex
    \begin{abstract}
      Parametric statistical models that are implicitly defined in
      terms of a stochastic data generating process are used in a wide
      range of scientific disciplines because they enable accurate
      modeling. However, learning the parameters from observed data is
      generally very difficult because their likelihood function is
      typically intractable. Likelihood-free Bayesian inference
      methods have been proposed which include the frameworks of
      approximate Bayesian computation (ABC), synthetic likelihood,
      and its recent generalization that performs likelihood-free
      inference by ratio estimation (LFIRE). A major difficulty in all
      these methods is choosing summary statistics that reduce
      the dimensionality of the data to facilitate
      inference. While several methods for choosing summary
      statistics have been proposed for ABC, the literature for
      synthetic likelihood and LFIRE is very thin to date. We here
      address this gap in the literature, focusing on the important
      special case of time-series models. We show that convolutional
      neural networks trained to predict the input parameters from the
      data provide suitable summary statistics for LFIRE. On a wide
      range of
      time-series models, a single neural network architecture
      produced equally or more accurate posteriors than alternative
      methods.
    \end{abstract}
    \section{Introduction}
    \label{sec:intro} 
    We consider the task of estimating the posterior density $\post$
    of parameters $\bt$ given observed data $\obs$ when the
    statistical model $p(\bx \mybar \bt)$ is implicitly specified in
    terms of a stochastic computer program that takes the model
    parameters $\bt$ as input and generates samples $\bx$ from $p(\bx
    \mybar \bt)$ as output. Such models enable accurate modeling of
    possibly nonlinear stochastic phenomena and are widely used in
    scientific disciplines as diverse as genetics \citep{Tavare1997,
      Arnold2018}, physics \citep{Cameron2012, Tietavainen2017},
    ecology and evolution \citep{Hartig2011, Corander2017},
    econometrics \citep{Gourieroux1996,Frazier2018}, and
    vision and robotics
    \citep{Mansinghka2013,LopezGuevara2017a}.

    While the computer program enables us to generate samples from
    $p(\bx \mybar \bt)$, it does not provide us with a direct way of
    evaluating $p(\bx \mybar \bt)$ --- the models are thus said to be
    implicitly defined \citep{Diggle1984}. Moreover, numerical
    evaluation of $p(\bx \mybar \bt)$ for all but the simplest
    implicit models is prohibitively expensive, which means that the
    likelihood function $L(\bt) = p(\obs \mybar \bt)$ is not available
    either and estimating the posterior becomes very difficult.

    Several likelihood-free Bayesian inference methods to estimate the
    posterior exist when only sampling from the model is possible. The
    methods include approximate Bayesian computation \citep[ABC,
    ][]{Tavare1997, Pritchard1999}, synthetic likelihood
    \citep{Wood2010, Price2017} and its generalizations
    \citep{Dutta2016, Fasiolo2018}; for recent reviews, see
    e.g.\ \citep{Lintusaari2017, Sisson2018}. The methods rely on
    summary statistics $\sumstats$ that reduce the dimensionality of
    the data. ABC uses the summary statistics to assess the similarity
    between the simulated and observed data (by typically computing
    the Euclidean distance between them), while the synthetic
    likelihood approach models the summary statistics as a Gaussian
    distribution for each parameter value; its generalizations relax
    the Gaussianity assumption.
    
    The summary statistics thus crucially affect the estimated
    posterior. In ABC, there has been considerable work on learning or
    selecting suitable summaries using dimensionality
    reduction methods and methods from regression and classification \citep[see
      e.g.\ ][]{Aeschbacher2012, Fearnhead2012, Blum2013, Gutmann2018,
      jiang_learning_2018}. But for the synthetic likelihood approach
    and its generalizations, the literature is very thin to
    date.

    To evaluate the synthetic likelihood pointwise at a value of
    $\bt$, we have to estimate and invert the covariance matrix of the
    summary statistics, which can pose numerical challenges. Robust
    methods have been proposed that can be considered to correspond to
    some form of summary statistics selection or transformation:
    \citet{Wood2010} proposes preconditioning and reweighing of the
    summary statistics, \citet{Ong2017} uses shrinkage estimation of
    the covariance matrix and \citet{An2018} the graphical lasso to
    obtain sparse estimates of its inverse. In the generalization of
    the synthetic likelihood by \citet{Dutta2016}, named
    likelihood-free inference by ratio estimation (LFIRE), the authors
    use their method to automatically select and combine relevant
    summary statistics from a larger pool of candidates. But much like
    the aforementioned approaches for the synthetic likelihood, it is
    assumed that the list of candidate summary statistics contains
    suitable ones in the first place.

    The aim of this article is to lift this burden on the user and to
    enhance LFIRE with a practical method that automatically learns
    suitable summary statistics from the raw data $\bx$. We propose
    that predicted parameter values, computed directly from the raw
    data, provide summaries that are well suited for LFIRE. Focusing
    on the special but important case of time-series data and
    stochastic dynamical models, we show that convolutional neural
    networks are particularly apt to learn such summary statistics for LFIRE.
 
    The rest of the article is structured as follows. Section
    \ref{sec:approach} explains the proposed approach and Section
    \ref{sec:toy-models} validates the method on models where the
    posterior can be accurately computed. In Section
    \ref{sec:realistic-models}, we apply it to complex models with
    intractable likelihoods, and Section \ref{sec:conclusions} concludes
    the paper.

    \section{Learning summary statistics for LFIRE on time-series data}
    \label{sec:approach} 
    We first review LFIRE and discuss which summary statistics are
    suitable for this likelihood-free inference framework. We then
    propose to learn them for time-series data by using convolutional
    neural networks and then present the proposed method.
    
    \subsection{Summary statistics for LFIRE}
    The LFIRE approach of \citet{Dutta2016} formulates the problem of
    posterior density estimation as a problem of estimating the ratio
    $r(\bt,\bx$) between the data generating distribution $p(\bx
    \mybar \bt)$ and the marginal $p(\bx) = \int p(\bx \mybar \bt)
    \prior \ud \bt$, where $\prior$ is the prior over
    $\bt$.\footnote{A related approach for estimating likelihood-ratios
      can be found in \citep{Cranmer2015}.} After the ratio is
    estimated, the posterior follows directly from Bayes' theorem,
    \begin{align}
       \label{eq:bayes}
       \post &= r(\bt,\obs) \prior, & r(\bt,\bx) & = \frac{p(\bx \mybar \bt)}{p(\bx)}.
    \end{align}
    For models specified by a data generating process, we cannot
    evaluate $p(\bx \mybar \bt)$ and $p(\bx)$ but we can sample from
    the two distributions.\footnote{To sample from the marginal
      $p(\bx)$, we first sample a $\bt'$ from the prior and then a
      $\bx$ from $p(\bx \mybar \bt')$.} This is exploited by
    \citet{Dutta2016} who estimate the ratio $r(\bt, \bx)$ by training
    a logistic regression model to learn to classify between data
    sampled from $p(\bx \mybar \bt)$ and $p(\bx)$ \citep[other methods
      to estimate the ratio can also be used,
      see][]{Gutmann2011b,Sugiyama2012b}.

    In more detail, let $\Xtheta = \{ \xtheta^{(i)} \}$ be a set of
    $\ntheta$ samples from $p(\bx \mybar \bt)$, $\Xm = \{ \xm^{(i)}
    \}$ a set of $\nm$ samples from $p(\bx)$, and $\hbeta(\bx)$ a
    parametric model for $\log r(\bt, \bx)$ for any given value of
    $\bt$. \citet{Dutta2016} learn the value of the log ratio at
    $\bt$ by minimizing the logistic loss
    \begin{align}
      \mathcal{L}_{\bt}(\bb) =& \frac{1}{n} \left\{
      \sum_{i=1}^{\ntheta} \log\left[1+\nu
        \exp(-\hbeta(\xtheta^{(i)})\right] + \right. \nonumber \\  &
      \left. \sum_{i=1}^{n_m} \log\left[1+\frac{1}{\nu}
        \exp(\hbeta(\xm^{(i)})\right] \right\} + \Omega(\bb)
      \label{eq:log-loss}
    \end{align}
    with respect to $\bb$, where $\Omega(\bb)$ is a regularizing
    penalty term.\footnote{After learning, $\betahat(\bt) =
      \text{argmin}_{\bb} \mathcal{L}_{\bt}(\bb)$ and
      $\hat{h}(\bt,\bx) = h_{\betahat(\bt)}(\bx) \approx \log
      r(\bt,\bx)$.} While other parametrizations of
    $\hbeta(\bx)$ are possible, for their empirical results,
    \citet{Dutta2016} worked with
    \begin{align}
      \hbeta(\bx) & = \bb^\top \sumstats(\bx)
      \label{eq:linear-model}
    \end{align}
    where $\sumstats(\bx)$ is a fixed vector-valued nonlinear
    transformation of the raw data $\bx$. They are the features in the
    classification problem and correspond to summary statistics in
    likelihood-free inference. \citet{Dutta2016} selected and combined
    the relevant ones by using the $L_1$ penalty $\Omega(\bb) =
    \lambda \sum_i |\beta_i|$ together with cross-validation to choose
    the penalty strength $\lambda$.

    The issue of choosing summary statistics in likelihood-free
    inference manifests itself in the LFIRE framework in the question
    of how to model $\log r(\bt,\bx)$, or equivalently, which features
    to choose for classification. Ideally, we would like to have
    features for which classification can be performed with a simple
    decision boundary in the feature space. Our approach to generate
    summary statistics for LFIRE thus consists in learning features
    for which the decision boundary has a particularly simple form.

    The main idea is that features (summary statistics) that are
    suitable for LFIRE can be learned by inverting the data process,
    that is, by learning to predict the value of $\bt$ from the data
    $\bx$. The learned predictors $\bthat(\bx)$ then define the
    desired summary statistics for LFIRE. Indeed, if the learning is
    done well, the predictions based on $\bx \sim p(\bx \mybar \bt)$
    will cluster around $\bt$ while the predictions based on $\bx \sim
    p(\bx)$ are spread out over the domain of the prior so that a
    simple elliptical decision boundary can be used to perform the
    classification. This means that we can work with a parametric
    model $\hbeta(\bx)$ as in \eqref{eq:linear-model} where
    $\sumstats(\bx)$ is given by the predicted parameter values
    $\hat{\theta}_i(\bx)$ as well as all their squares and unique
    pair-wise combinations and a constant, i.e. for a $d$-dimensional
    parameter space
    \begin{align}
      \sumstats = (\hat{\theta}_1, \ldots, \hat{\theta}_d,
      \hat{\theta}_1^2, \hat{\theta}_1\hat{\theta}_2, \ldots,
      \hat{\theta}_{d-1}\hat{\theta}_d,\hat{\theta}_d^2, 1)
      \label{eq:pairwise-sumstats}
    \end{align}
    where we suppressed the dependency on $\bx$. Figure \ref{fig:ARCH_features}
    illustrates this idea for the ARCH model (see below).
    \begin{figure}[t]
     \centering
     \includegraphics[trim={0 0.5cm 0 1.35cm},clip,
       width=0.35\textwidth]{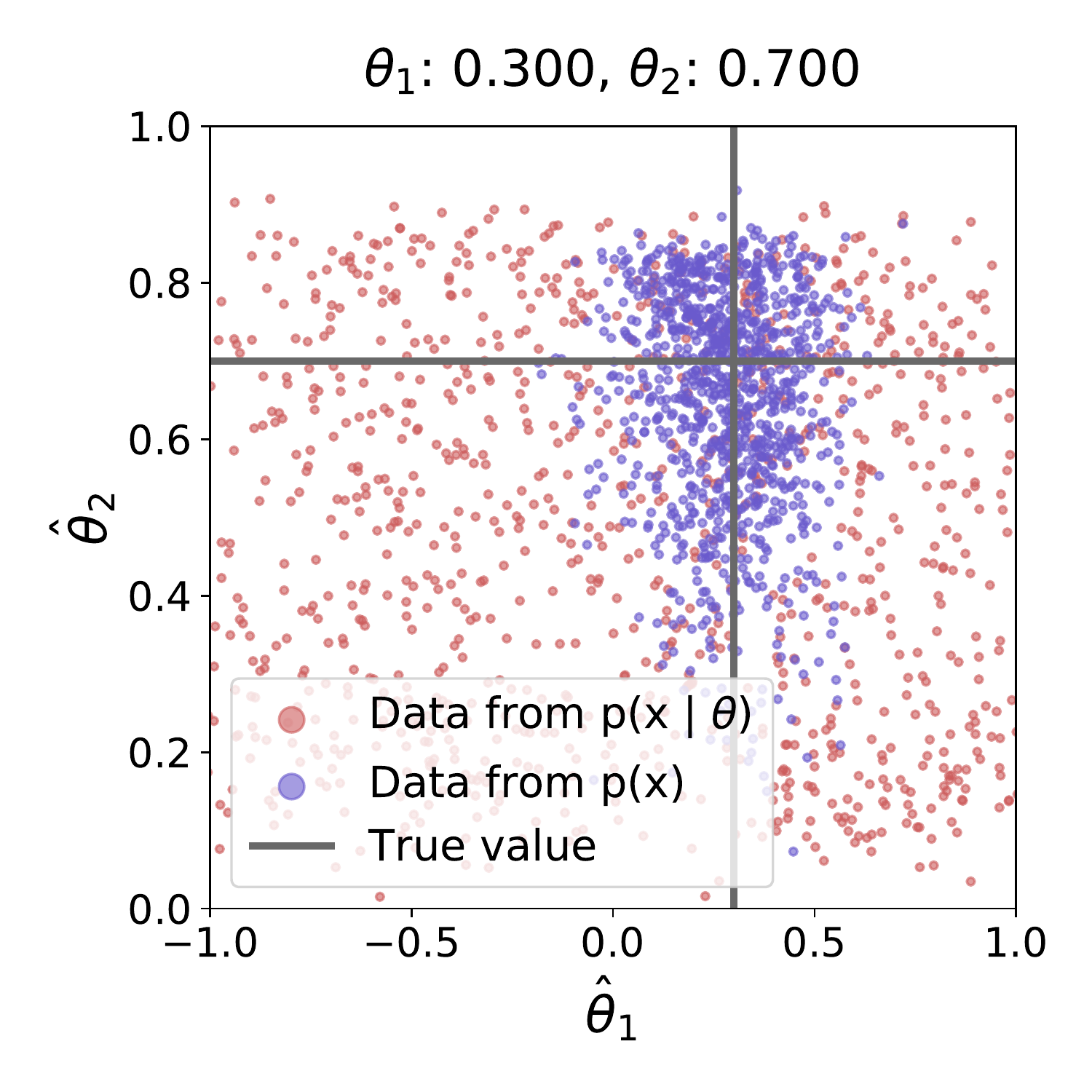}
        \caption{\label{fig:ARCH_features} We propose to use
          predictors $\bthat(\bx)$ to define the summary statistics
          for LFIRE. For $\bx \sim p(\bx \mybar \bt)$, here $\bt =
          (0.3, 0.7)$, the predictions are concentrated around $\bt$
          (blue) while for $\bx \sim p(\bx)$, the predictions are
          spread out in the parameter space (red). An elliptic
          decision boundary can be used to classify the two data
          clouds and hence to learn the ratio $r(\bt,\bx)$ in
          \eqref{eq:bayes}.}
    \end{figure}
    \vspace{-1ex}
    \subsection{Learning summary statistics for time-series data}
    \vspace{-0.5ex}
    The predictors $\bthat(\bx)$ can be learned from parameter-data
    pairs $(\bt,\bx)$ with $\bx \sim p(\bx \mybar \bt)$ by (nonlinear)
    regression, and neural networks provide a very flexible function
    class among which to search for the predictors. Importantly,
    however, no single network architecture and training method will
    work for all kinds of data. Since we would like to learn the
    predictors with as little user guidance and manual tuning as
    possible, we focus on time-series data where the
    popular convolutional networks provide a restricted yet flexible
    enough function class.

    Convolutional networks have fewer parameters to learn and
    are suited for describing time-series data because they can
    capture higher-order statistical dependencies between different
    time points. Related (nonlinear) autocorrelation functions have
    been used as summary statistics in previous work
    \citep[e.g.][]{Wood2010}.
    We will see that a \emph{single} generic neural network architecture is
    able to produce suitable summary statistics for LFIRE on a range
    of different time-series models and data sets.

    Neural networks have been used before in ABC: \citet{Blum2010,
      Papamakarios2016} used them in the context of regression ABC
    where the summary statistics are assumed to be given, and, more
    relevant for the topic of this paper, \citet{jiang_learning_2018}
    used them to learn summary statistics from raw data for use in
    ABC. The main difference of the latter work to this paper is (a)
    the different likelihood-free inference framework --- ABC versus
    LFIRE and (b) the focus on time-series and the use of convolutional
    neural networks.

    \vspace{-2.5ex}
    \subsection{Proposed method} \vspace{-1ex}
    \label{sec:method}
    In line with the above, the proposed method to enhance LFIRE with
    summary statistics for time-series data separates into two distinct
    stages. The first is to train a convolutional neural network to
    predict $\bt$ from $\bx$. The second is to run
    LFIRE with the model $\hbeta(\bx)$ for the log-ratio in
    \eqref{eq:linear-model} and summary statistics defined by the
    learned predictors $\bthat(\bx)$ as in
    \eqref{eq:pairwise-sumstats}.
    
    The training data $(\bt_i,\bx_i)$, $i=1, \ldots, m$, for the first
    stage is obtained by sampling $\bt_i$ from the prior $\prior$ and
    time-series data $\bx_i$ from the implicitly defined $p(\bx \mybar
    \bt_i)$. Throughout all experiments in this paper, $m = 100,000$
    and the data were split into $80,000$ training and $20,000$
    validation examples.

    In all simulations, we used the \emph{same} training procedure and
    architecture for the convolutional neural network (called the
    ``DireNet'' --- ``dire'' to indicated that we perform
    likelihood-free inference for dynamical models). The neural
    network consisted of two convolutional layers, the first followed by a
    max-pooling layer, and then a fully connected layer to capture
    long-range dependencies. We used rectified linear units (ReLu) as
    activation functions for these layers. ReLu activation functions
    are preferred over sigmoidal or hyperbolic tangent activation
    functions since they are less susceptible to vanishing gradients
    when training the neural network
    \citep[e.g.][]{hochreiter_vanishing_1998}. For $d$ dimensional
    model parameters $\bt$, the output layer consisted of $d$ output
    units, where we used the linear activation function. The detailed
    neural network architecture and training procedure is presented in
    the supplementary material.
    
    The LFIRE objective in \eqref{eq:log-loss} was minimized using the
    R package glmnet \citep{friedman_regularization_2010} as done
    before by \citet{Dutta2016}. We also used the same settings as
    them: We used $1,000$ data points from $p(\bx)$ and $p(\bx \mybar
    \bt)$, and 10-fold cross-validation to select the $L_1$
    regularization strength $\lambda$.

    LFIRE yields a surrogate posterior over the parameters $\bt$
    \citep{Dutta2016}. If needed, samples from the posterior can be
    obtained by using it as the target distribution in any
    sampler. In this paper, however, sampling from the posterior was
    not necessary. The parameters $\bt$ are low-dimensional (as often
    the case in likelihood-free inference problems) so that posterior
    expectations were simply computed by taking weighted sums over a
    grid.

    \vspace{-1ex}
    \section{Inference for toy models}       \vspace{-1ex}
    \label{sec:toy-models}
    We illustrate and validate the proposed method on two toy models,
    and compare it to LFIRE with expert (manual) summary statistics and
    summary statistics defined by the deep network of
    \citet{jiang_learning_2018}.

    \subsection{Models}
    \label{sec:toy-model-description}
    The first model considered is the autoregressive conditional
    heteroscedasticity (ARCH) model defined by
    \begin{align}
   \hspace{-2ex} x^{(t)} &= \theta_1 x^{(t-1)}  + e^{(t)}, & e^{(t)}  &= \zeta^{(t)}  \sqrt{\alpha +
              \theta_2 (e^{(t-1)})^2}
    \end{align}
    with $t = 1, \dots ,100$, $x^{(0)} = 0$, $\alpha = 0.2$, and where
    $\zeta^{(t)} $ and $e_0$ are independent standard normal random
    variables. The parameters of interest are $\bt =
    (\theta_1,\theta_2)$ for which we assume a uniform prior on $[-1,
      1] \times [0, 1]$. By forward simulating the above equations, we
    can easily sample time-series data $\bx = (x^{(1)} , \ldots,
    x^{(100)})$ from the model. The exact posterior can be computed
    numerically to high accuracy \citep[see e.g.\ ][Supplementary Material
      1.2.4]{Gutmann2018}.

    The second model considered is the moving average model of
    order two (MA2) which is described by
    \begin{align}
        x^{(t)} &= e^{(t)}  + \theta_1 e^{(t-1)}  + \theta_2 e^{(t - 2)}, & x^{(0)}  &= e^{(0)} 
    \end{align}
    where $t=1, \ldots, 100$ and the $e^{(t)}$ are independent
    standard normal variables. The parameters of interest are $\bt =
    (\theta_1, \theta_2)$. We used as prior the uniform distribution
    on the triangle defined by $\theta_1 \in [-2, 2]$, $\theta_1 +
    \theta_2 > -1$, and $\theta_1 - \theta_2 < 1$ as
    \citet{Marin2012}. We can again generate time-series data
    $\bx = (x^{(1)} , \ldots, x^{(100)})$ from the model and the
    exact posterior can be computed numerically to high accuracy
    (see supplementary material).
    \subsection{Results}
    \begin{traizq}{ARCH model. Comparing the prediction accuracy of the DireNet and the deep net by \citet{jiang_learning_2018}. For $R^2$, larger is better; for MSE, smaller is better. The test set had size $100,000$.}{arch-reconstruction}{}
  Measure     & DireNet         & Jiang et al \\ \hline
  Train $R^2$ & \textbf{0.834}  & 0.782     \\ 
  Test $R^2$   & \textbf{0.827}  & 0.740     \\
  Train MSE   & \textbf{0.019} & 0.028 \\
  Test MSE     & \textbf{0.020} & 0.032 \\
    \end{traizq}
    
  We first assess how well the DireNet can reconstruct the parameters
  $\bt$ from the raw data $\bx$, that is how good the learned
  predictors $\bthat(\bx)$ are. We compare the reconstruction results
  with those for the deep neural network by
  \citet{jiang_learning_2018} in terms of the mean-squared error
  and the coefficient of determination ($R^2$),
\begin{align}
  R^2 & = 1 - \frac{\sum_i ||\bthat(\bx_i)  - \bt_i||_2^2}{\sum_i ||\bt_i - \bar{\bt}||_2^2},
\end{align}
where $\bt_i \sim \prior$ and $\bx_i \sim p(\bx \mybar \bt_i)$. Table
\ref{tab:arch-reconstruction} shows that the proposed DireNet obtains
a higher $R^2$ value and a lower MSE on both the training and the
test set, the latter of which indicating better generalization
performance. It also shows that the gap between the training and the
test performance is smaller for the DireNet, which indicates
less overfitting. The reduction in overfitting is to be expected given
the reduced number of parameters in the DireNet ($8,422$ vs.\ $30,502$),
but the better generalization and training performance indicates that
using convolutional layers is beneficial for the time-series data.

Figure \ref{fig:ARCH_features} shows example reconstructions
$\bthat(\bx)$ obtained by the DireNet. Further examples and
corresponding plots for the deep network by
\citet{jiang_learning_2018} are shown in the supplementary
material. In the DireNet figures, the reconstructions for $\bx \sim
p(\bx \mybar \bt)$ are clustered around $\bt$ (in blue) while the
reconstructions for $\bx \sim p(\bx)$ (in red) are spread out over the
domain of the prior as desired. We note that perfect reconstructions
are not strictly necessary for LFIRE to work. This is because training
of the classifier in LFIRE can accommodate systematic biases or
distortions in the predictions $\bthat(\bx)$; for LFIRE to work, we
only need that (sets of) predictions $\bthat(\bx)$ for different $\bt$
are distinguishable from one another (see supplementary material).
   
\begin{traizq}{KL divergences between the exact posterior and the one learned by LFIRE with summary statistics given by the DireNet, the deep net by \citet{jiang_learning_2018}, and manual expert statistics. Averages $\pm$ standard errors for 500 inference tasks are shown.}{toy_all}{}
  Model & DireNet & Jiang et al & Manual \\
  \hline
  ARCH &  \textbf{0.481} \scriptsize{$\pm$ 0.017} &  0.959 \scriptsize{$\pm$ 0.025 } &  0.751 \scriptsize{$\pm$ 0.046} \\
  MA2 &  \textbf{0.842} \scriptsize{$\pm$ 0.025 } &  1.631 \scriptsize{$\pm$ 0.040} &  1.384 \scriptsize{$\pm$ 0.041} \\
\end{traizq}

We next compare the accuracy of the inferred posteriors when using the
DireNet, the deep network by \cite{jiang_learning_2018}, and manual
expert summary statistics in LFIRE. For the ARCH model, the expert
statistics are auto-correlations and auto-covariances with lag up to
order five as in \citep{Dutta2016}. For the MA2 model, we used
auto-correlations with lag up to two as \citet{Marin2012}. For all
summary statistics, we included their pairwise combinations and a
constant as features in LFIRE, so that all methods use the same model
for the log-ratio. For all methods, LFIRE was used with a $L_1$
penalty that prunes away unnecessary features.

For both models, we sampled $500$ parameters from the prior $\prior$
and generated for each an observed data set $\obs$, for which we then
inferred the posterior $p(\bt \mybar \obs)$. Since the exact posterior
can be computed for both models (see above), we assessed the accuracy
of the learned posterior by the Kullback-Leibler (KL) divergence
between them. For its computation, we used a $20\times20$ rectangular
grid spanning the support of the prior.\footnote{The grid was the same for all methods and the numerical values
  of the KL divergences are reported up to the (common) stepsize.}

The results are summarized in \autoref{tab:toy_all}. For both models,
we see a significant improvement when using the DireNet as compared to
both other approaches. Furthermore, in this case, the manually chosen
statistics lead to better performance than the deep
network  when used in LFIRE. Example posteriors for the ARCH model are shown in Figure
\ref{fig:arch_posterior}. Further examples are provided in the supplementary
material.

\begin{figure*}[t!]
  \centering
  \begin{subfigure}[t]{0.29\textwidth}
    \centering
            \includegraphics[width=\textwidth]{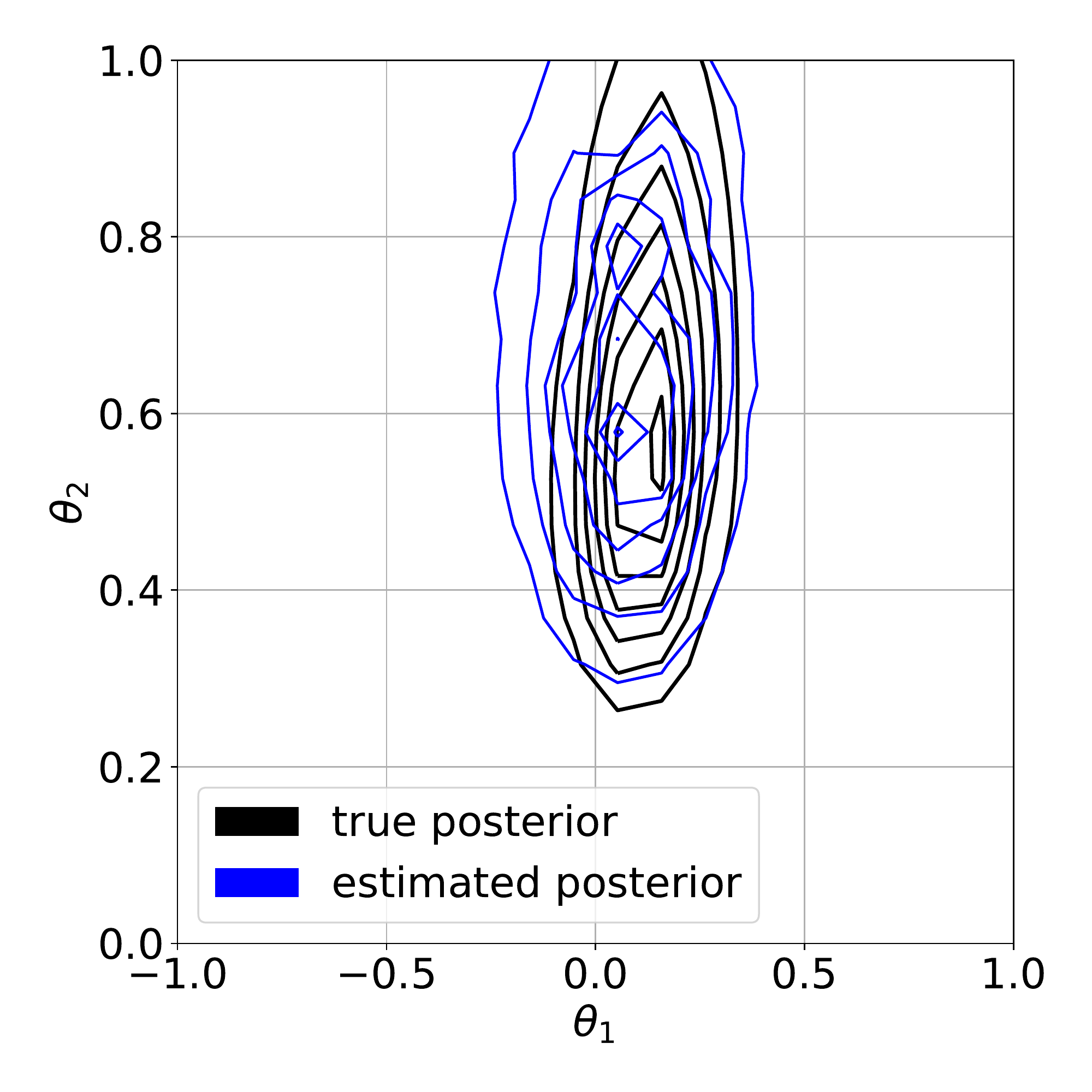}
            \caption{DireNet}
            \label{fig:arch_dire}
  \end{subfigure}
  \hspace{1ex}
  \begin{subfigure}[t]{0.29\textwidth}
    \centering
    \includegraphics[width=\textwidth]{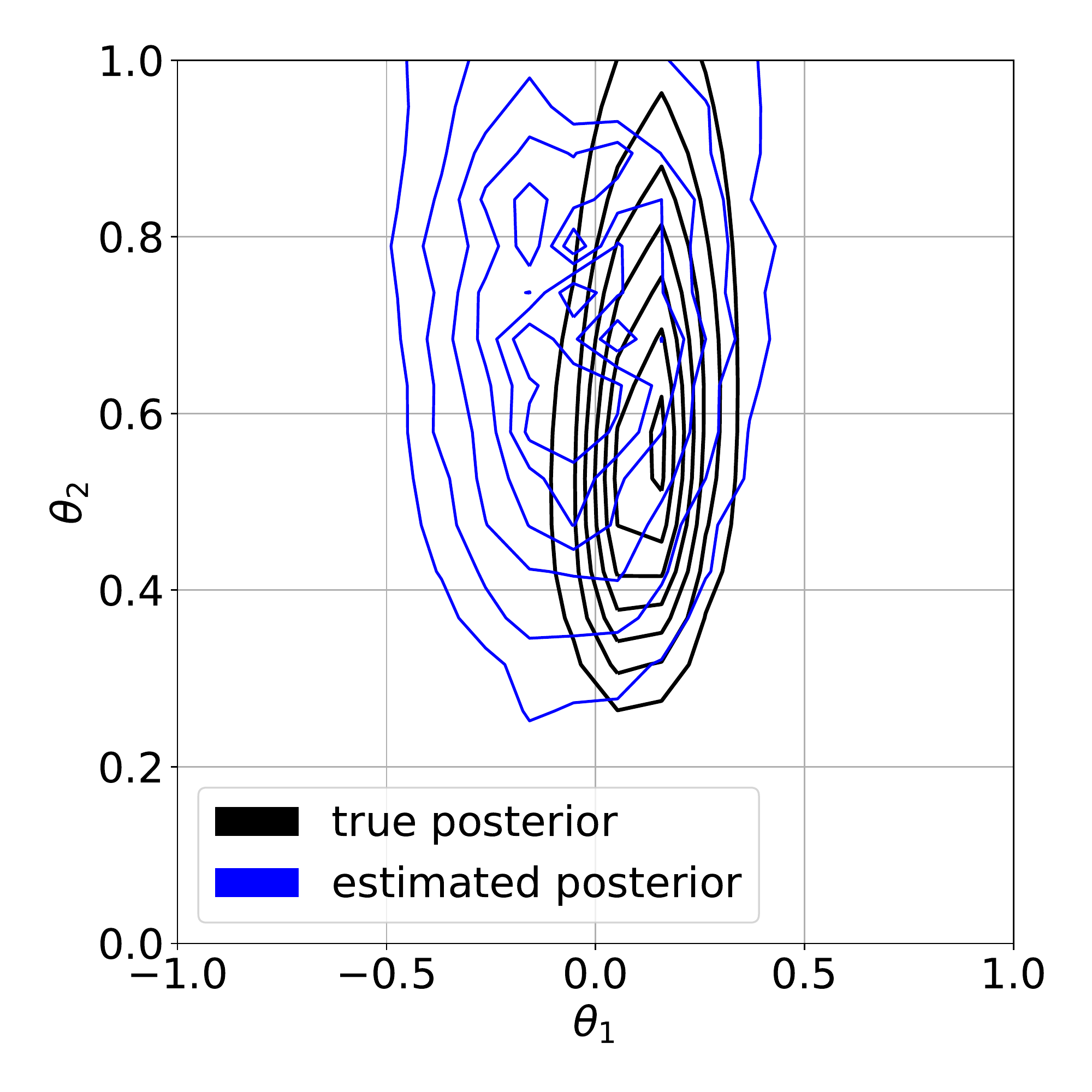}
    \caption{Deep net by \textcite{jiang_learning_2018}}
    \label{fig:arch_jiang}
  \end{subfigure} 
  \hspace{1ex}
  \begin{subfigure}[t]{0.29\textwidth}
    \centering
    \includegraphics[width=\textwidth]{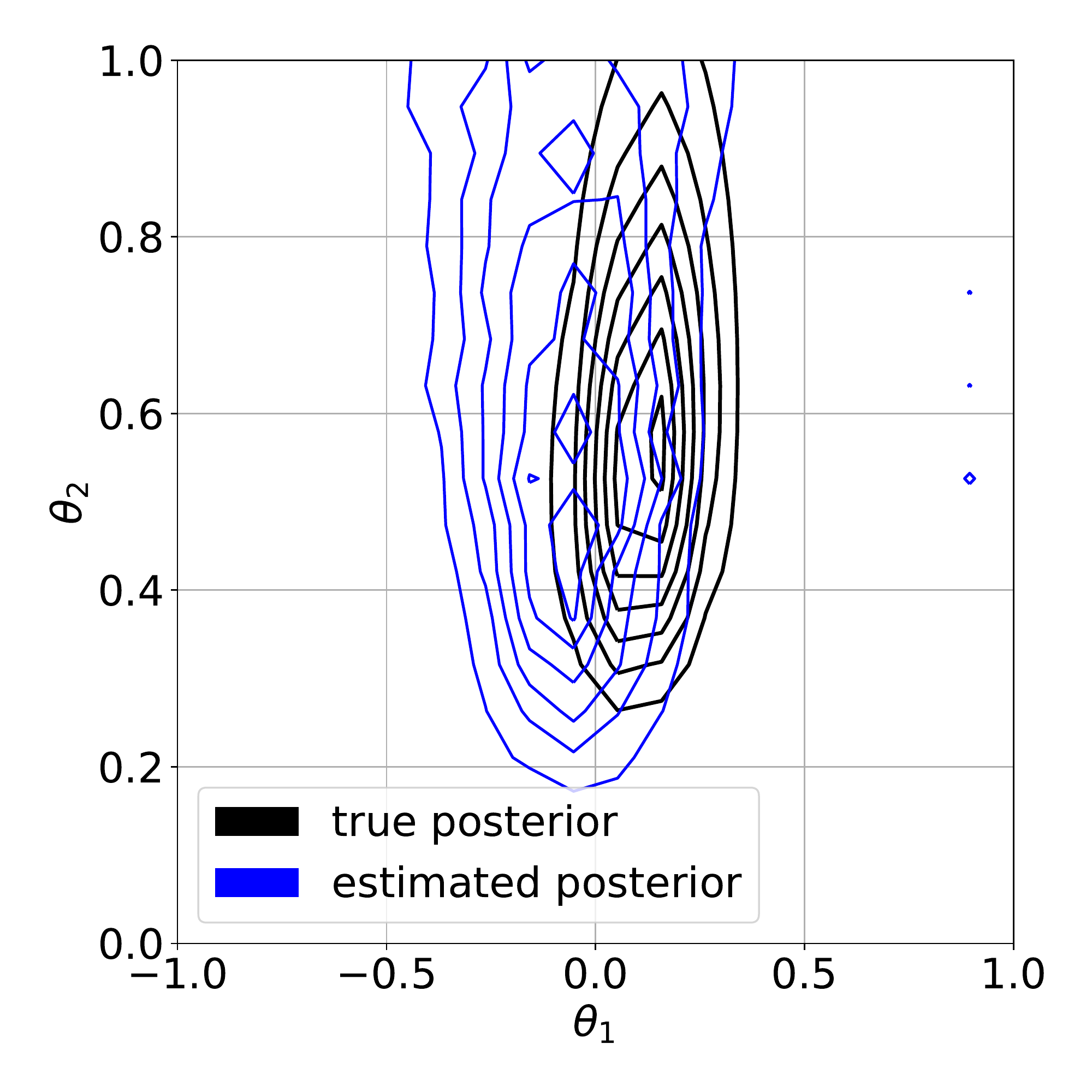}
    \caption{Manual summary statistics}
    \label{fig:arch_manual}
  \end{subfigure} 
  \caption{ARCH model: example posteriors for $\obs$ generated with $\bt = (0.3, 0.7)$. Posteriors estimated by LFIRE are in blue, the exact (true) posterior is in black. (a-c) differ in the summary statistics used for LFIRE.}
  \label{fig:arch_posterior}
\end{figure*}

\vspace{-1ex}
    \section{Inference for complex models}\vspace{-1ex}
    \label{sec:realistic-models}
    We here apply our method to real-world models with intractable
    likelihoods and compare its performance to alternative methods.
    \vspace{-1ex}
    \subsection{Lotka-Volterra model}
    The Lotka-Volterra model is a continuous-time Markov chain that
    can be used to model predator-prey dynamics in ecology and
    chemical reactions \citep{boys_bayesian_2008}. The generative process for the case of two species
    $(x_1^{(t)},x_2^{(t)})$ is defined by the transition distribution
    \begin{align*}
        p(&x_1^{(t + \delta t)} = a_1, x_2^{(t + \delta t)} = a_2\ |\  
            x_1^{(t)} = b_1, x_2^{(t)} = b_2) = \\
        &\begin{cases}
            1 - \gamma^{(t)}\delta t + o(\delta t) &
                \text{if}\ a_1 = b_1\ \text{and}\ a_2 = b_2 \\
            \theta_1 b_1\delta t  + o(\delta t) & \text{if}\ a_1 = b_1 + 1\ \text{and}\ a_2 = b_2 \\
            \theta_2 b_1 b_2 \delta t + o(\delta t) &
                \text{if}\ a_1 = b_1 - 1\ \text{and}\ a_2 = b_2 + 1 \\
            \theta_3 b_2  \delta t+ o(\delta t) &
                \text{if}\ a_1 = b_1\ \text{and}\ a_2 = b_2 - 1 \\
            o(\delta t) & \text{otherwise}
        \end{cases}
    \end{align*}
    where $\gamma^{(t)} = \theta_1 b_1 + \theta_2 b_1 b_2 + \theta_3
    b_2$. The model has three parameters $\bt = (\theta_1, \theta_2,
    \theta_3)$ and for each we assume a uniform prior: $\theta_1 \in
          [e^{-2}, 1]$, $\theta_2 \in [e^{-5}, e^{-2.5}]$, and
          $\theta_3 \in [e^{-2}, 0]$.  We can sample from the model
          with the algorithm by \citet{Gillespie1977}, see also
          \citet{Fearnhead2012}, and in our simulations, we use
          time-series $\bx = (\ (x_1^{(1)},x_2^{(1)}), \ldots,
          (x_1^{(50)},x_2^{(50)})\ )$ of length fifty.

    We compared the performance of the DireNet to LFIRE with manual
    summary statistics. The manual summary statistics were the
    $x_1^{(t)}$ and $x_2^{(t)}$ as in previous work where ABC was used
    for inference \citep[e.g.][]{Toni2009}.\footnote{With the
      quadratic expansion in LFIRE as before.} For the comparison, we
    generated 500 observed data sets by sampling ``true'' data
    generating parameters $\bt$ from the prior $\prior$ and then
    solved the 500 inference problems. Since the likelihood is
    intractable, we assessed the performance by the relative error
    between the estimated posterior means and the true data generating
    parameters as done in previous work \citep{Dutta2016}. We compute
    the relative error for each dimension $i$ of $\bt$, i.e.\
    \begin{equation}
      \label{eq:rel-error}
        \mathcal{RE}_i = \frac{|\mathbb{E}(\theta_i \mybar \obs) - \theta_i^{\text{true}}|} {|\theta_i^{\text{true}}|}.
    \end{equation}
    The expectation was approximated by a weighted sum over a
    $20\times20\times20$ rectangular grid covering the support of the
    prior.

    Using the same observed data sets for both the DireNet and the
    manual summary statistics allowed us to perform a point-wise
    comparison between the two methods. Following \citet{Dutta2016},
    we used the 500 relative errors of both methods to estimate the
    distribution of their difference $\Delta_i^\text{rel}$,
    \begin{equation}
      \label{eq:rel-error-diff}
      \Delta^\text{rel}_i = \mathcal{RE}_i^\text{DireNet} - \mathcal{RE}_i^\text{Manual}.
    \end{equation}
    If the distribution is skewed to the left, the proposed approach
    performs better and vice-versa. Similarly, negative expected values of 
    $\Delta_i^\text{rel}$ indicate better performance of the proposed method.
    
    Figure \ref{fig:lv_rel_error} shows the distribution for the three
    parameters. All distributions are skewed to the left and the
    (bootstrap) 95\% confidence intervals for the means are on the
    negative axis too. Together they indicate that the DireNet
    performs better than the manual statistics at estimating the
    parameters. The numerical values of the confidence intervals and
    further results are provided in the supplementary material.
   
    \subsection{Ricker model}
    \label{sec:ricker}
    The Ricker model \citep{ricker_stock_1954} describes the observed
    size of an animal population over time. The dynamics of the
    population size is modeled as     %
    \begin{equation*}
        \log N^{(t + 1)} = \log r + \log N^{(t)}  -N^{(t)} + \sigma e^{(t)}
    \end{equation*}
    where the $e_t$ are independent standard normal random
    variables. We assume that we cannot observe the true population
    size but only a noisy measurement modeled as a sample from a
    Poisson distribution with mean $\lambda_t = \phi N_t$ at time-steps
    $ t = 1, \dots ,T$ \citep{Wood2010}, with $T=50$. The noisy
    observations are our data $\obs$ and we would like to infer the
    log growth rate $\log{(r)}$, the noise standard deviation
    $\sigma$, and the scaling parameter $\phi$ of the observation
    model. As in previous work \citep{Wood2010}, we assume uniform
    priors: $\log{(r)} \in [3, 5]$, $\sigma \in [0, 0.6]$ and $\phi
    \in [5, 15]$.

    We compared our results with two other (non-LFIRE) inference
    methods --- synthetic likelihood with the summary statistics by
    \citet{Wood2010} and semi-automatic ABC by \citet{Fearnhead2012}. The comparison is done in the same way as
    before for the Lotka-Volterra model (using again 500 inference
    tasks).

    For the synthetic likelihood, we used the summary statistics and
    code provided by \citet{Wood2010}. \citet{Wood2010} used fourteen
    summary statistics that included the coefficients of the
    autocorrelation function and the coefficients of fitted nonlinear
    autoregressive models \citep[see the supplementary material of
    ][]{Wood2010}. For the
    computation of the mean and covariance matrix that is needed in
    the synthetic likelihood approach, we used $1,000$ model
    simulations. The posterior was then obtained by Markov Chain Monte
    Carlo using the same setup as \citet{Wood2010} with the exception
    that we reduced the variances of the proposal distribution to
    $0.02, 0.01$ and $0.05$ for $\log{(r)}$, $\log{(\sigma)}$ and
    $\log{(\phi)}$, respectively, following \citet{Gutmann2016a} who
    observed better mixing with those values.

    For the semi-automatic approach by \citet{Fearnhead2012}, we used
    the code kindly provided by the authors. In brief, this approach
    transforms the summary statistics by \citet{Wood2010} as well
    additional ones specified by the authors (the set ``E2'') and then
    performs ABC by Markov chain Monte Carlo using the transformed
    summary statistics. 
  
    For the proposed method with LFIRE, we computed the posterior mean
    using a weighted sum over a $20\times20\times20$ grid covering the
    prior while for the synthetic likelihood and semi-automatic ABC,
    it was computed by averaging the posterior samples.

\begin{figure*}[t!]
        \centering
        \begin{subfigure}[t]{0.32\textwidth}
            \centering
            \includegraphics[width=\textwidth]{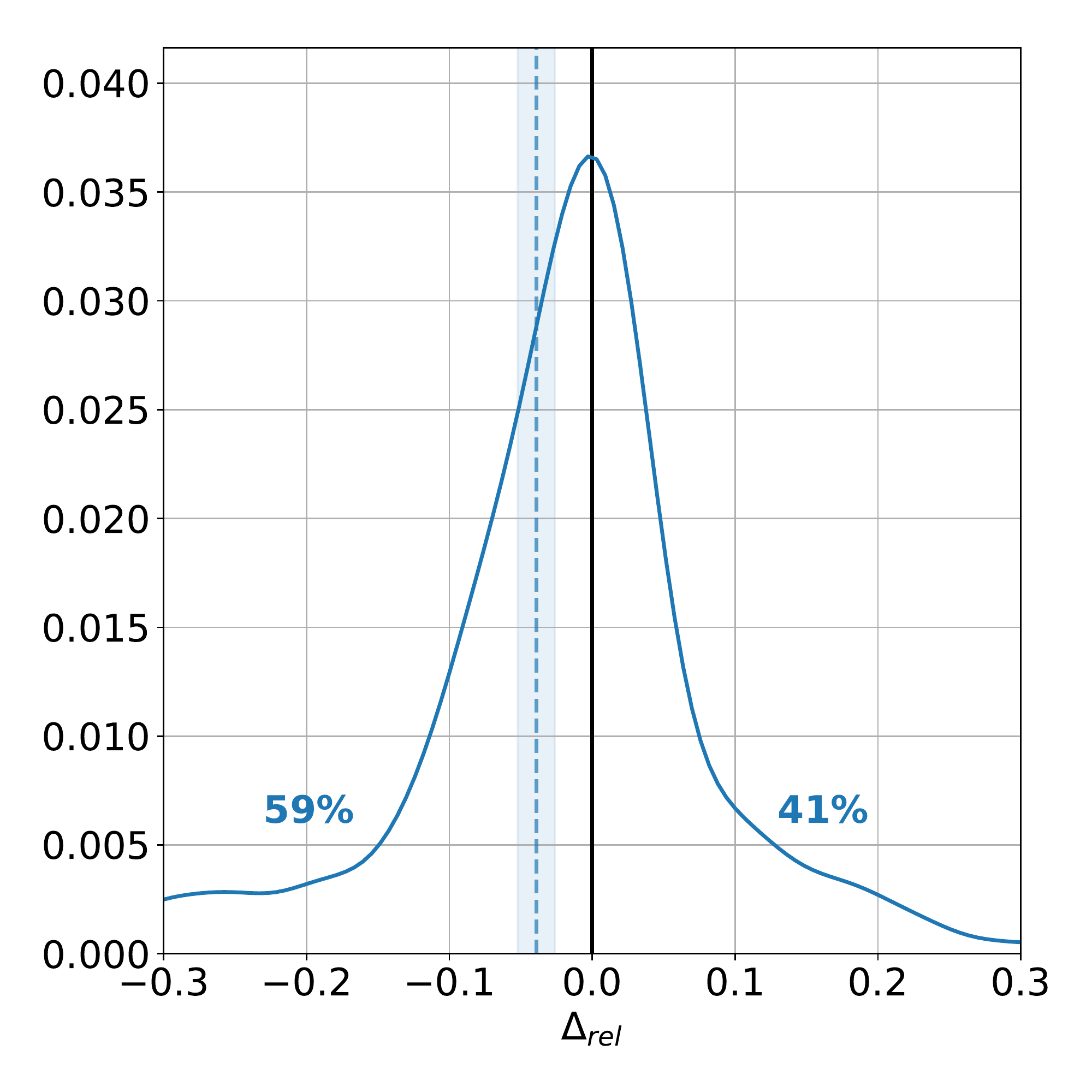}
            \caption{Distribution for $\theta_1$}
            \label{fig:lv_rel_error_theta1}
        \end{subfigure} \hfill
        \begin{subfigure}[t]{0.32\textwidth}
            \centering
            \includegraphics[width=\textwidth]{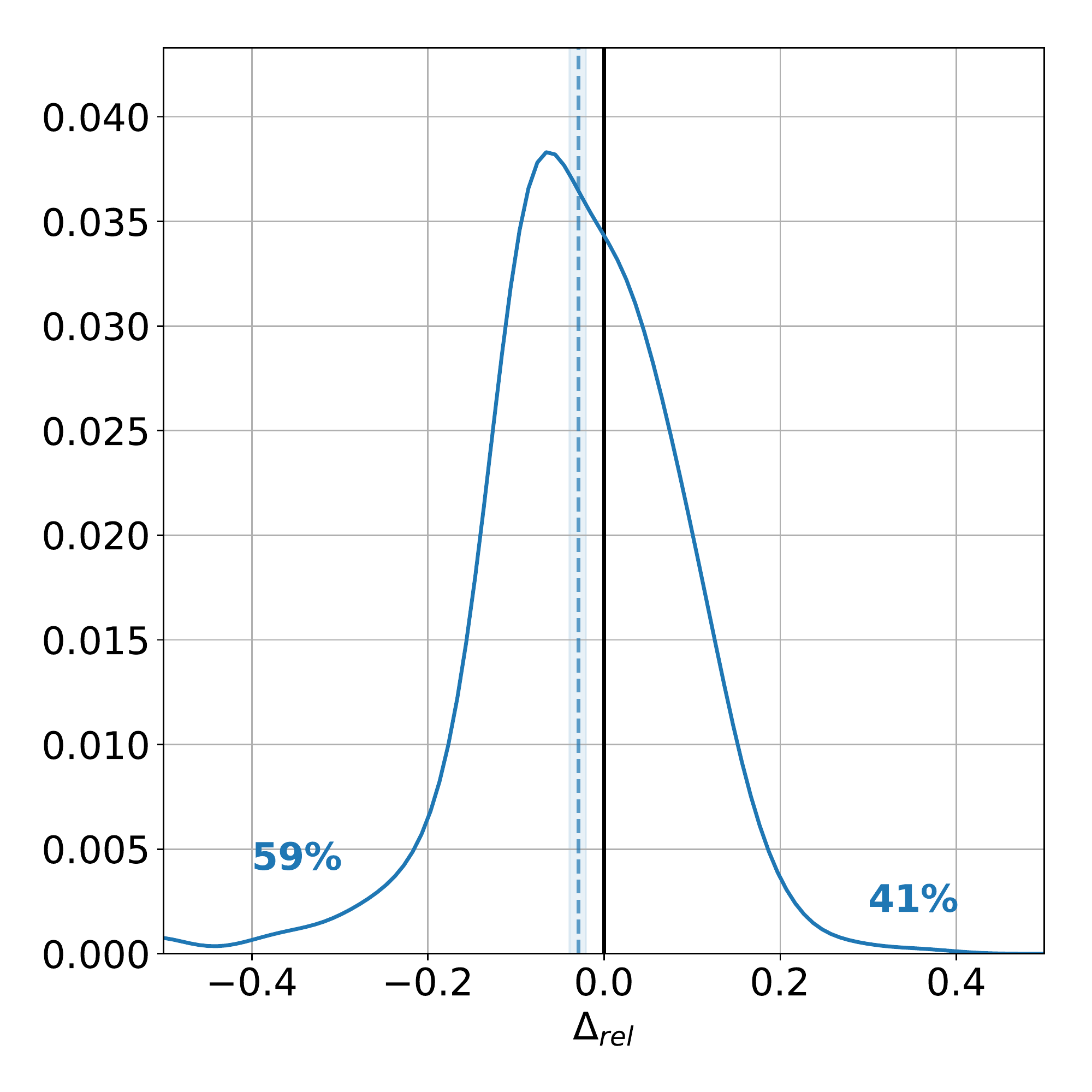}
            \caption{Distribution for $\theta_2$}
            \label{fig:lv_rel_error_theta2}
        \end{subfigure} \hfill
        \begin{subfigure}[t]{0.32\textwidth}
            \centering
            \includegraphics[width=\textwidth]{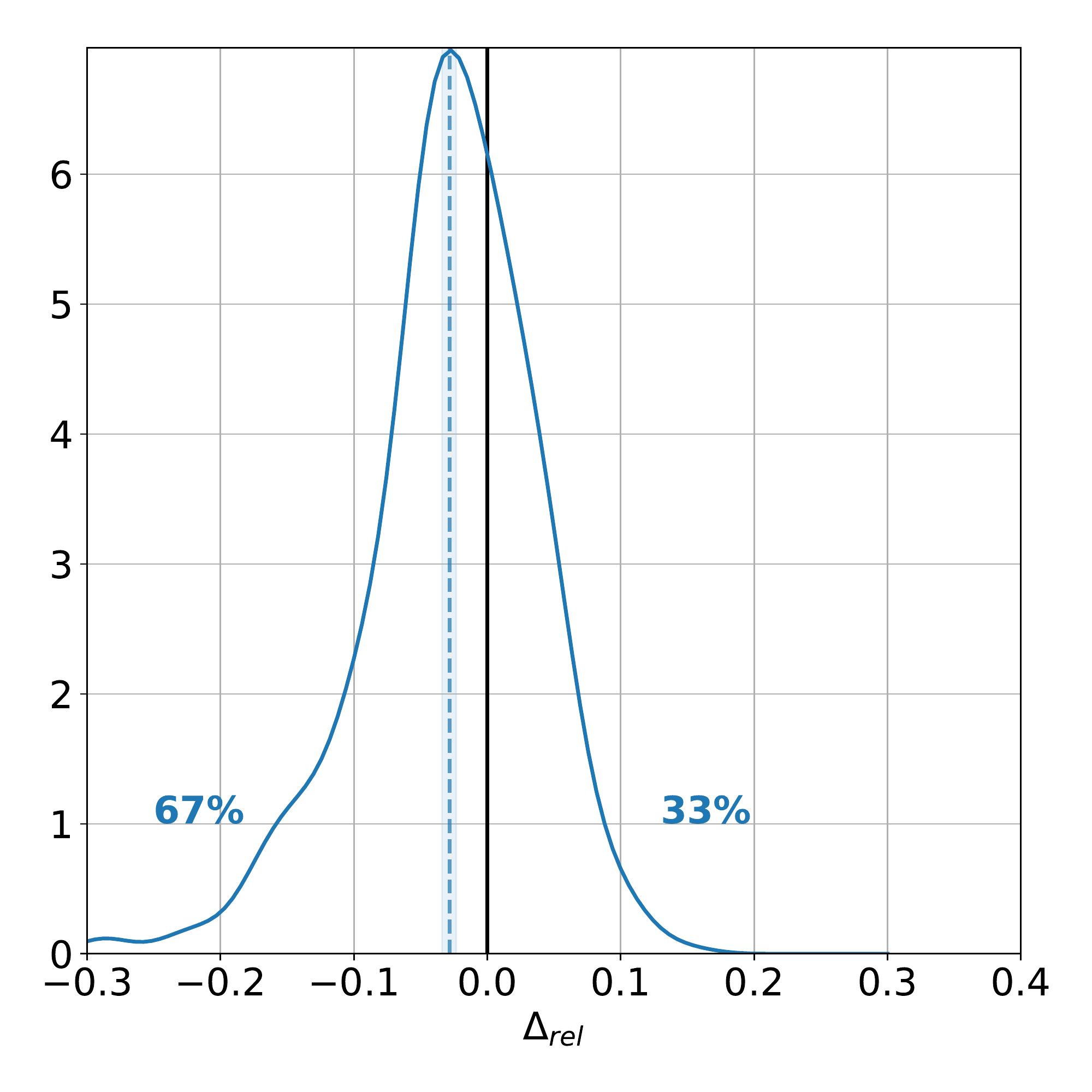}
            \caption{Distribution for $\theta_3$}
            \label{fig:lv_rel_error_theta3}
        \end{subfigure}
        \caption{Lotka-Volterra model: Distribution of the difference
          between the relative error for the DireNet and manual
          summary statistics. Negative values correspond to better performance of the proposed method. Bootstrap 95\% confidence intervals and
          mean are shaded in blue. The distributions are visualized as
          kernel density estimates with a Gaussian kernel (bandwidths
          for $\theta_1$ to $\theta_3$: 0.025, 0.04, and 0.02).}
        \label{fig:lv_rel_error}
    \end{figure*}

\begin{figure*}[t!]
        \centering
        \begin{subfigure}[t]{0.32\textwidth}
            \centering
            \includegraphics[width=\textwidth]{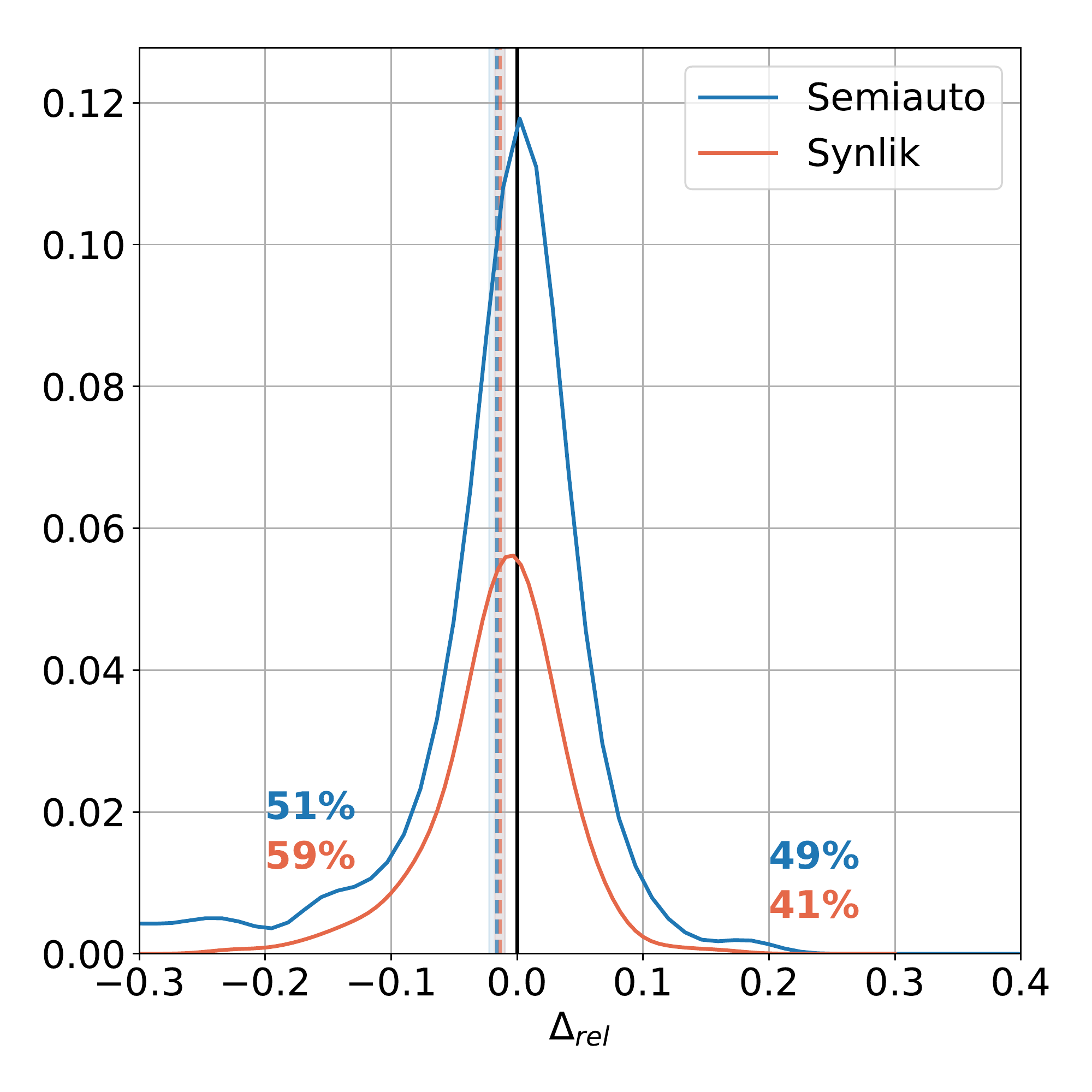}
            \caption{$\log{r}$}
            \label{fig:ricker_rel_error_logr_overlay}
        \end{subfigure}
        \hfill
        \begin{subfigure}[t]{0.32\textwidth}
            \centering
            \includegraphics[width=\textwidth]{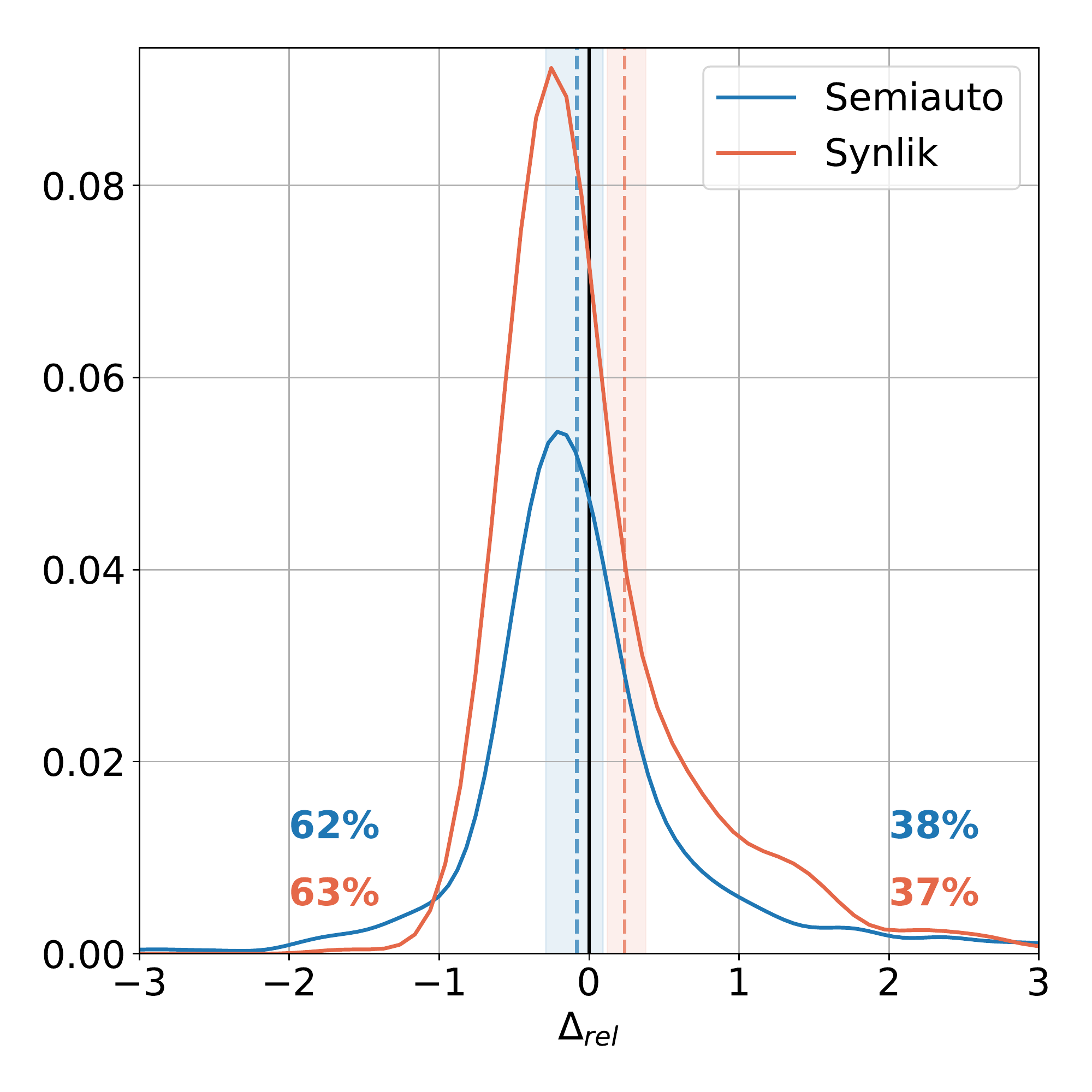}
            \caption{$\sigma$}
            \label{fig:ricker_rel_error_sigma_overlay}
        \end{subfigure}
        \hfill
        \begin{subfigure}[t]{0.32\textwidth}
            \centering
            \includegraphics[width=\textwidth]{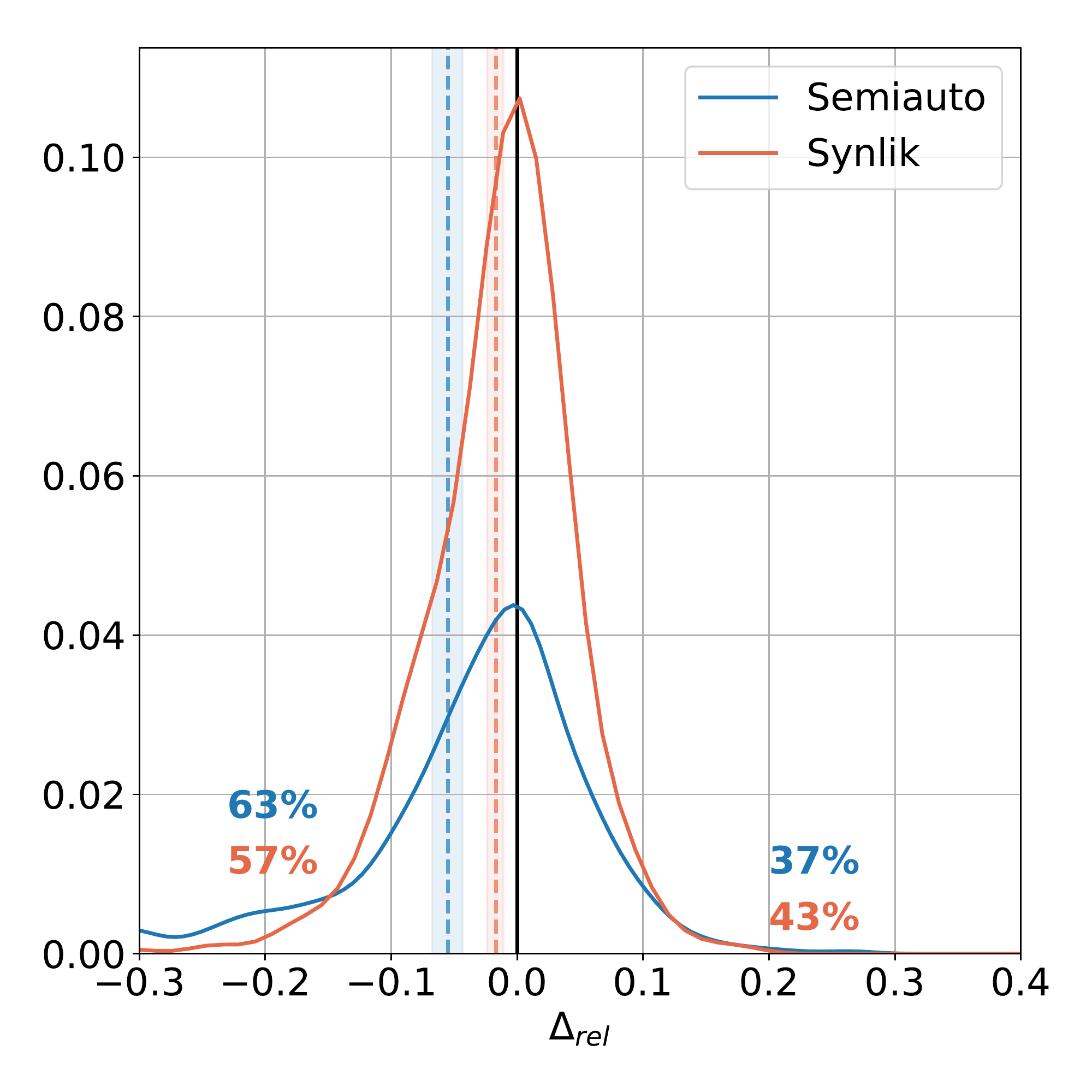}
            \caption{$\phi$}
            \label{fig:ricker_rel_error_phi_overlay}
        \end{subfigure}
        \caption{Ricker model: Distribution of the difference between
          the relative error for the DireNet and synthetic likelihood
          (red) and the DireNet and semi-automatic ABC
          (blue). Shown as above, bandwidths: 0.02, 0.2, 0.02.}
        \label{fig:ricker_rel_error_overlay}
    \end{figure*}

    Figure \ref{fig:ricker_rel_error_overlay} shows the distributions of
    the difference in the relative errors comparing our approach to
    synthetic likelihood and semi-automatic ABC in the same way as in Figure
    \ref{fig:lv_rel_error}. As before, distributions skewed to the left and
    negative expected values indicate better performance of the
    proposed method. We can see that for $\log{r}$ and $\phi$, LFIRE
    with DireNet summary statistics leads to more accurate parameter
    estimates (posterior means) than the other two approaches. For
    $\sigma$, the results are not conclusive: The most likely outcome
    (mode of the distribution) is that the proposed method yields a
    more accurate estimate but bootstrap 95\% confidence intervals on
    the mean include zero in case of the comparison to semi-automatic
    and are on the positive axis in case of the comparison to
    synthetic likelihood. On the other hand, the 95\% confidence
    intervals for the median are on the negative axis (see supplementary
    material). All in all, the differences for $\sigma$ are not
    clear-cut, which may not be surprising given that $\sigma$ is a
    difficult parameter to estimate (see e.g.\ the supplementary material).
    
    \subsection{Lorenz model}
    The Lorenz model of \citet{wilks_effects_2005} is a stochastic
    forty-dimensional time-series model for weather variables
    $x_k^{(t)}$, $k = 1 \dots 40$. These follow a system of coupled
    stochastic differential equations:
    \begin{align}
        \frac{\partial x_k^{(t)}}{\partial t} =& -x_{k - 1}^{(t)}
        (x_{k - 2}^{(t)} - x_{k + 1}^{(t)}) - x_k^{(t)} + F - \nonumber \\
        &g(x_k^{(t)}, \bt) + \eta_k^{(t)}
        \\ g(x_k^{(t)}, \bt) &= \theta_1 + \theta_2 x_k^{(t)}
    \end{align}
    where $F$ is a constant set to 10 and $\eta_k^{(t)}$ a stochastic
    perturbation term representing unobserved variables of faster time scales \citep[see
    ][]{wilks_effects_2005, Dutta2016}. For negative $k$ the model is
    cyclic, for example for $k = 1$, $k - 1 = 40$. We follow the cited
    previous work and numerically solve the above equations for $t \in
    [0,4]$ using a 4th order Runge-Kutta solver discretizing the time
    interval into $160$ time-steps, each of $\Delta t = 0.025$.

    As in previous work, all initial values $x_k^{(0)}$ are assumed
    known, and the parameters of interest are $\bt =
    (\theta_1,\theta_2)$. The prior for $\bt$ was the uniform
    distribution on $[0.5, 3.5] \times [0, 0.3]$.

    We used exactly the same DireNet as in all the other
    inference tasks before, thus taking only temporal convolutions. Since the time
    series is forty-dimensional, one could also also have used
    spatio-temporal convolutions rather than just temporal ones
    further reducing the number of neural network parameters.

    We compared the performance of the proposed method to LFIRE with
    the six manual summary statistics from \citet{Dutta2016}, which
    were the mean; variance and auto-covariance with lag one of the
    $x_k^{(t)}$ variables; the cross covariance with lag one
    of $x_k^{(t)}$ with $x_{k-1}^{(t)}$; and $x_{k+1}^{(t)}$, all averaged
    over all dimensions. The setup and analysis was as before for the
    Lotka-Volterra and Ricker model (using again 500 inference tasks).
    
\begin{figure*}[t!]
        \centering
        \begin{subfigure}[t]{0.32\textwidth}
            \centering
            \includegraphics[width=\textwidth]{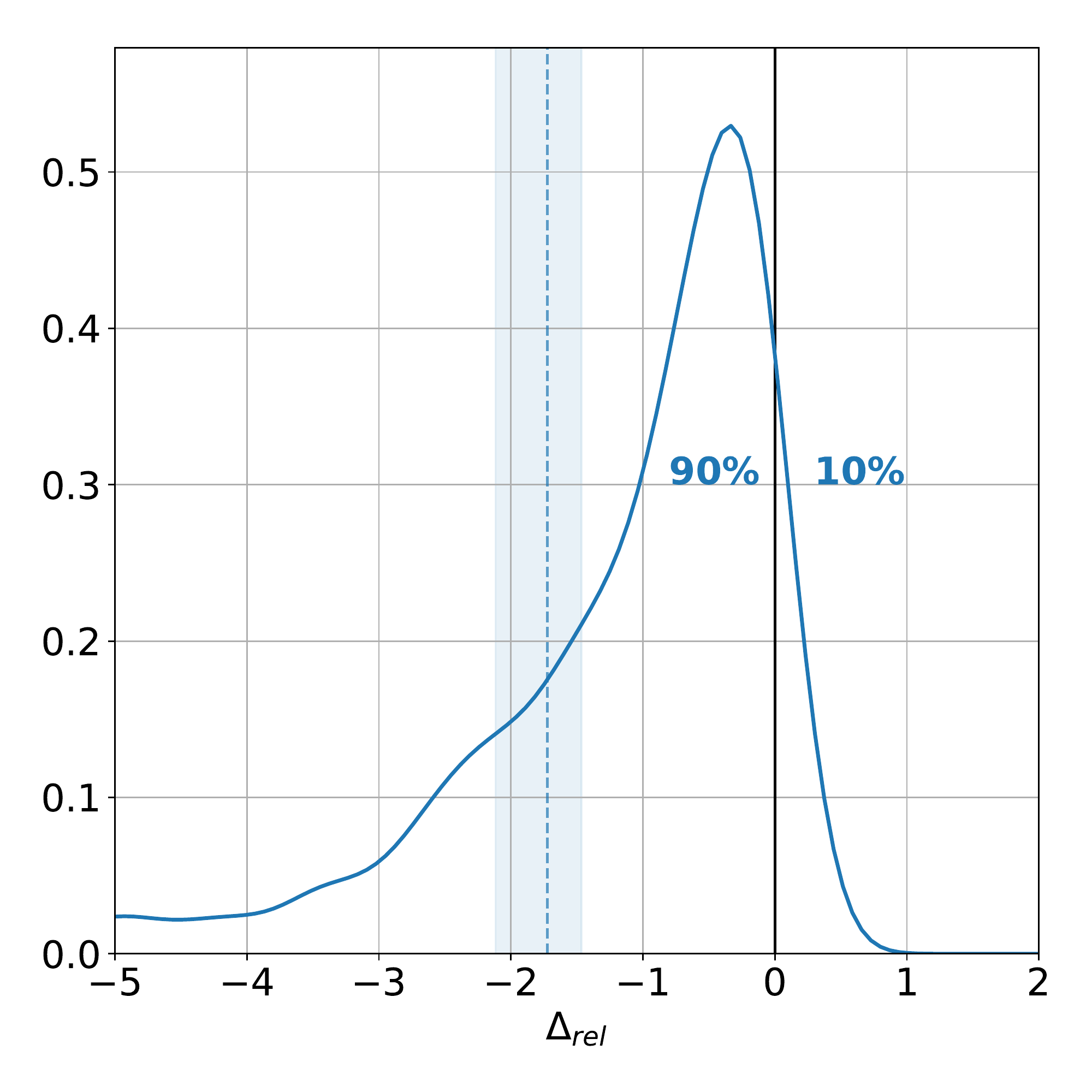}
            \caption{Relative error for $\theta_1$}
            \label{fig:lorenz_rel_error_t1}
        \end{subfigure}
        \hspace{4ex}
        \begin{subfigure}[t]{0.32\textwidth}
            \centering
            \includegraphics[width=\textwidth]{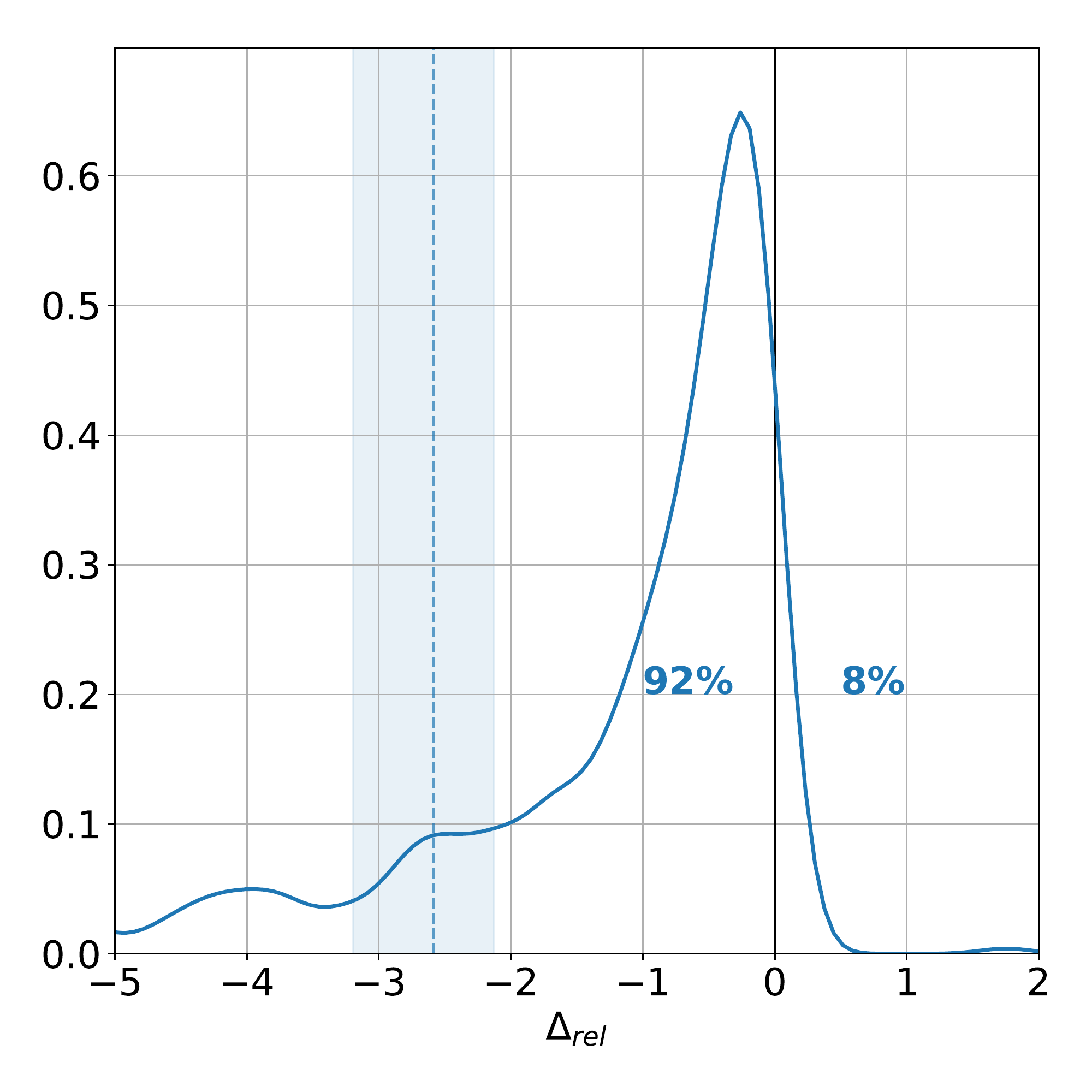}
            \caption{Relative error for $\theta_2$}
            \label{fig:lorenz_rel_error_t2}
        \end{subfigure}
        \caption{Lorenz model: Distribution of the difference between
            the relative error for the DireNet and manual summary
            statistics. Visualized as in Figure
            \ref{fig:lv_rel_error}, bandwidths: 0.3 and 0.2. \label{fig:lorenz_rel_error}}
    \end{figure*}

    The results are shown in Figure \ref{fig:lorenz_rel_error}
    (additional results with the numerical values for the bootstrap
    intervals, scatter plots showing true vs learned parameters, and
    example posteriors are in the supplementary material). There is a
    massive improvement over the manual summary statistics for both
    parameters. Indeed our approach outperforms here the manual summary
    statistics by the biggest margin, which may be due to the
    fact that time and experience of the research community has not
    yet optimized the manual summary statistics for this model in
    contrast to the Lotka-Volterra and Ricker model.


    \section{Conclusions}
    \label{sec:conclusions}
    We considered the problem of learning summary statistics
    for likelihood-free inference by ratio estimation \citep[LFIRE,
    ][]{Dutta2016}, an approach that generalizes the synthetic
    likelihood method by \citet{Wood2010, Price2017}. We proposed to
    use parameter values predicted from raw data as summary statistics
    for LFIRE. Focusing on the important case of dynamical models and
    time-series data, we showed that convolutional neural networks are
    well suited to learn such summary statistics. On a wide range of
    different models, a \emph{single} generic neural network
    architecture produced posterior estimates that are equally good or
    better than alternative, typically more customized, inference
    methods.

    While we focused on time-series models, neural networks will
    likely provide suitable summary statistics for other large model
    classes too. For example, convolutional networks should be useful
    for data with spatial or spatio-temporal structure. Moreover,
    since LFIRE generalizes synthetic likelihood, our findings should
    also be useful for that approach.
  
    The convolutional network only needs to be trained once per model
    and can then be deployed in arbitrarily many inference tasks
    (amortized inference). However, LFIRE requires solving an
    optimization problem for each parameter value for which we
    evaluate the posterior. While the outcome of the optimization can
    be stored and re-used for inference with different observed data
    sets, the repeated optimization is computationally
    costly. Importantly, however, this is an issue related to the
    current state of LFIRE and is not related to the method proposed
    in this paper. Techniques discussed and proposed in
    \citep{Dutta2016} and \citep{Cranmer2015} can be used to amend or
    alleviate this shortcoming, e.g.\ via transfer learning and
    Bayesian optimization \citep{Gutmann2016a}, and the proposed
    method can seamlessly be combined with these kinds of improvements.

    Overall our results suggest that modern techniques for training
    neural networks are promising for likelihood-free Bayesian inference, and
    we expect that further drawing upon techniques from artificial
    intelligence will lead to further advances in this challenging area
    of statistics.

%% file: supplementary.tex
    \section{Neural networks}
    In total, we performed simulations with three different neural
    networks. In addition to the convolutional DireNet and the deep
    network by \textcite{jiang_learning_2018} mentioned in the main
    text, we also performed simulations with a deep network similar to
    the one by \textcite{jiang_learning_2018} but trained with modern
    techniques including Batch Normalization \citep{ioffe_batch_2015}
    and Dropout \citep{srivastava_dropout:_2014}. This was done to
    have an additional baseline for comparison to the DireNet. In the
    following section we explore in more detail the effects of using
    each of those networks.

\begin{figure}[p!]
        \centering
        \begin{subfigure}[t]{0.32\textwidth}
            \centering
            \includegraphics[width=.8\textwidth]{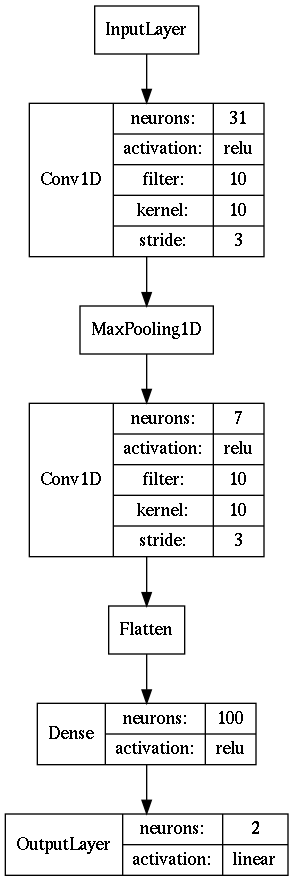}
            \caption{DireNet (convolutional network)}
            \label{fig:direnet}
        \end{subfigure}
        \begin{subfigure}[t]{0.32\textwidth}
            \centering
            \includegraphics[width=.8\textwidth]{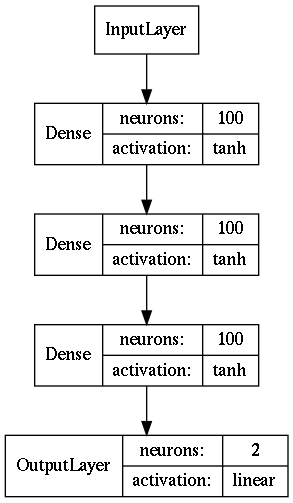}
            \caption{The network by \textcite{jiang_learning_2018}}
            \label{fig:jign_network}
        \end{subfigure}
        \begin{subfigure}[t]{0.32\textwidth}
            \centering
            \includegraphics[width=.8\textwidth]{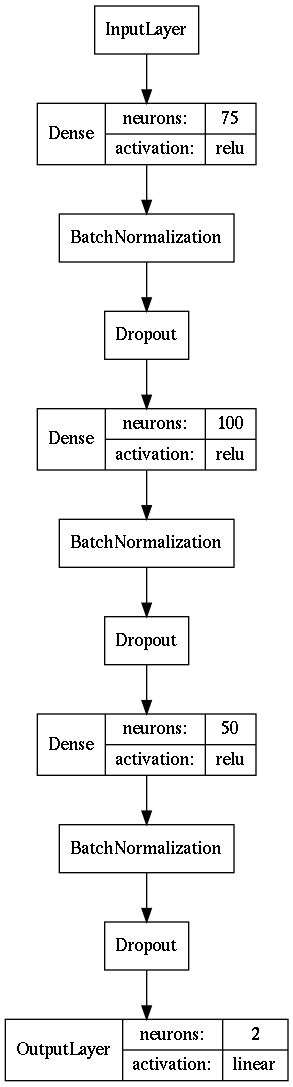}
            \caption{Additional deep neural network}
            \label{fig:deepnet}
        \end{subfigure}
        \caption[Neural network variants.]{Neural networks
          used. Compared to the network used in
          \cite{jiang_learning_2018}, the additional deep network that
          we considered has rectified linear activations and was
          trained with batch normalization and dropout. We also
          reduced the number of parameters by having less neurons in
          each hidden layer.\label{fig:direnets}}
    \end{figure}

    The three networks are shown in Figure \ref{fig:direnets}. The
    output layers have either $2$ or $3$ neurons, depending on the
    model in question. For instance, for the ARCH model, we have two
    output parameters $\bt = (\theta_1, \theta_2)$ and hence $2$
    neurons in the final layer. Excluding the output layer, the
    convolutional DireNet has $8,220$ parameters, the deep network by
    \textcite{jiang_learning_2018} $30,300$, and the additional deep
    network $21,125$ parameters. For the case of two outputs, the
    total number of parameters for the three networks become $8,422$;
    $30,502$; and $21,227$.
       
    Convolutional architectures have several hyperparameters per
    layer. For the convolutional layers, the \textit{kernel} parameter
    is the size of the convolutional kernel, the \textit{stride}
    parameter specifies how many units the kernel is shifted after
    each computation. We also have a number of these kernels,
    specified by the parameter \textit{filter}. After the
    convolutional layers we have a \textit{Flatten} layer that
    re-arranges the outputs so that fully-connected or \textit{Dense}
    layers can accept them.

    Throughout the paper we used $80,000$ samples for training and
    $20,000$ samples for validation. We trained the networks with the
    Adam optimizer \citep{Kingma2015} for $100$ epochs using a batch
    size of $256$ and early stopping with a patience of $30$ epochs
    \citep{prechelt_early_2012}. The deep network uses batch
    normalization \citep{ioffe_batch_2015} and dropout for the hidden
    layers with standard dropout rate $0.5$
    \citep{srivastava_dropout:_2014}. For DireNet, we used $L_2$
    regularization on the output layer with penalty strength set to
    the same value that \citet{jiang_learning_2018} used (namely
    $0.001$).

  \section{Posterior for the moving average model}
  This section contains a derivation of the posterior distribution for
  the moving average model of order 2 (MA2). The derivation follows
  that of \citet[Supplementary Material 1.2.4]{Gutmann2018} with the
  addition of a second parameter, so that the model is parametrized by $\bt = (\theta_1,\theta_2)$.

  The data generated by the model is a time-series $\bx_{1:\text{T}} = (x^{(1)}, \ldots,x^{(T)})$, which is defined as follows:
    \begin{align}
        x^{(t)} &= \eps^{(t)} + \theta_1 \eps^{(t - 1)} +
            \theta_2 \eps^{(t - 2)}, &
        x^{(0)} &= \eps^{(0)}, &
        x^{(1)} &= \eps_1 + \theta_1 \eps_0.
    \end{align}
    The $\eps_t$ are standard normal variables. If we consider the (column) vector $\beps = (\eps_0, \dots, \eps_t)$ of all such variables, then $\bx = \B \beps$, where:
    \begin{equation}
        \B = \begin{pmatrix}
                1 & 0 & 0 & 0 & \dots & 0\\
                \theta_1 & 1 & 0 & 0 & \dots & 0 \\
                \theta_2 & \theta_1 & 1 & 0 & \dots & 0 \\
                & & \ddots \\
                0 & 0 & \dots & \theta_2 & \theta_1 & 1 
            \end{pmatrix}
    \end{equation}
    We then know that $\beps \sim \mathcal{N}(\bf{0}, \mathbb{I})$ by definition above, where $\mathbb{I}$ denotes the identity matrix. Hence:
    \begin{equation}
        \bx_{0:\text{T}} \sim p(\bx_{0:\text{T}}\ |\ \bt)
            = \mathcal{N}(\bf{0}, \B\B^\top)
    \end{equation}
    However, we do not observe $x^{(0)}$. Therefore we define the matrix $\M$, which the matrix $\B \B^\top$ with the first row and column removed. We thus know that:
    
    \begin{equation}
        \bx_{1:\text{T}} \sim p(\bx_{1:\text{T}}\ |\ \bt)
            = \mathcal{N}(\bf{0}, M)
    \end{equation}
     
    By using the definition of conditional probability, we obtain the posterior distribution of $\bt$,
    \begin{align}
        p(\bt\ |\ \bx_{1:\text{T}}) =
            \frac{
                p(\bx_{1:\text{T}}\ |\ \bt)
            }
            {
                p(\bx_{1:\text{T}})
            } =
            \frac{
                p(\bx_{1:\text{T}}\ |\ \bt)
            }
            {
                \int_{\bt} p(\bx_{1:\text{T}}\ |\ \bt)\ \prior \ \text{d}\bt
            },
    \end{align}
where $\prior$ denotes the prior of $\bt$. 
    
    \section{Supplementary results}

    \subsection{ARCH model}
    For the ARCH model, we computed parameter predictions (reconstructions)
    $\bthat(\bx)$ and posterior distributions for the three different
    neural network architectures mentioned above and for observed data generated with
    different values of $\bt$.

\begin{figure}
        \centering
        \includegraphics[width=\textwidth]{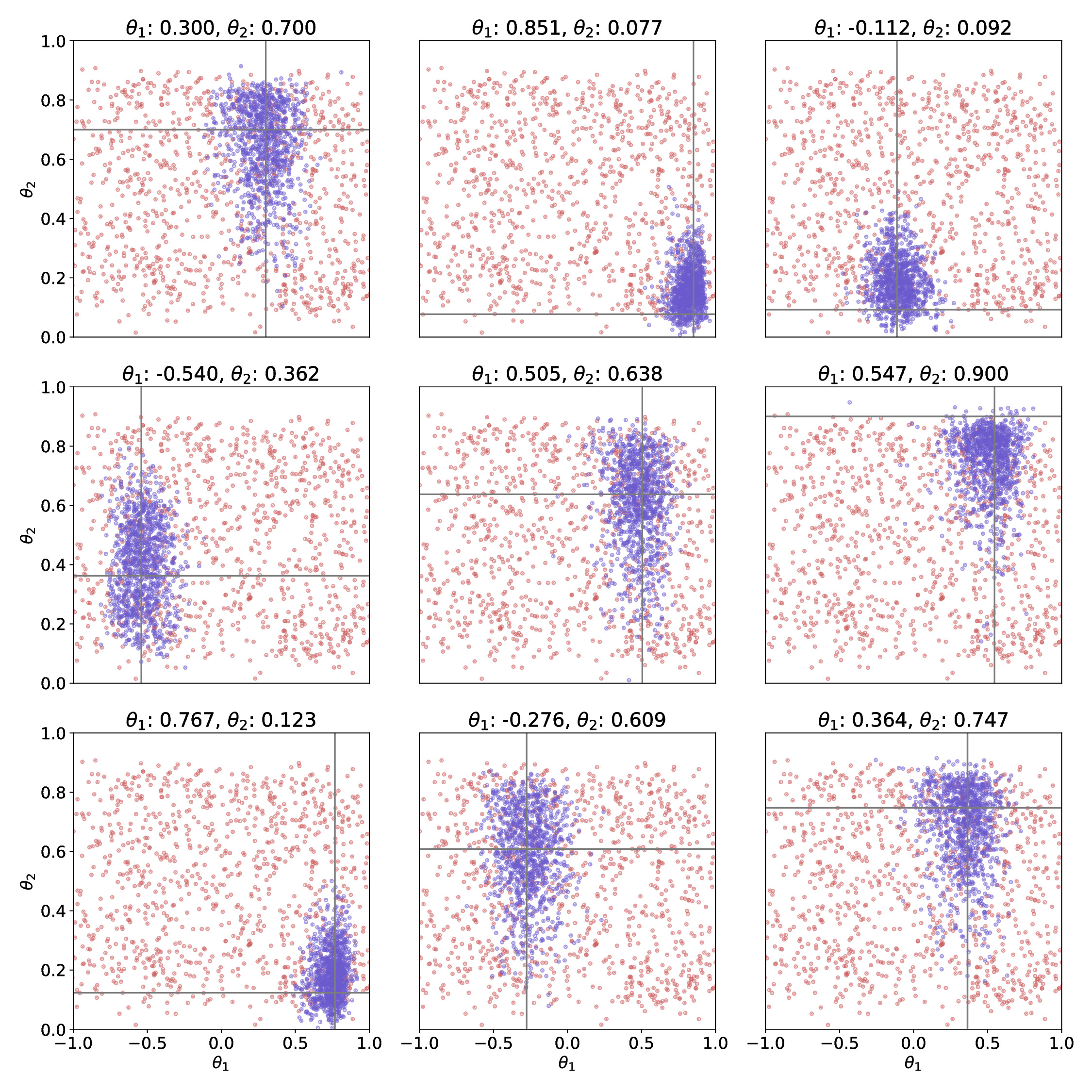}

        \caption{ARCH model: predictions $\bthat(\bx)$ by the DireNet}
        \label{fig:arch_scatterplot_direnet}
    \end{figure}

\begin{figure}
        \centering
        \includegraphics[width=\textwidth]{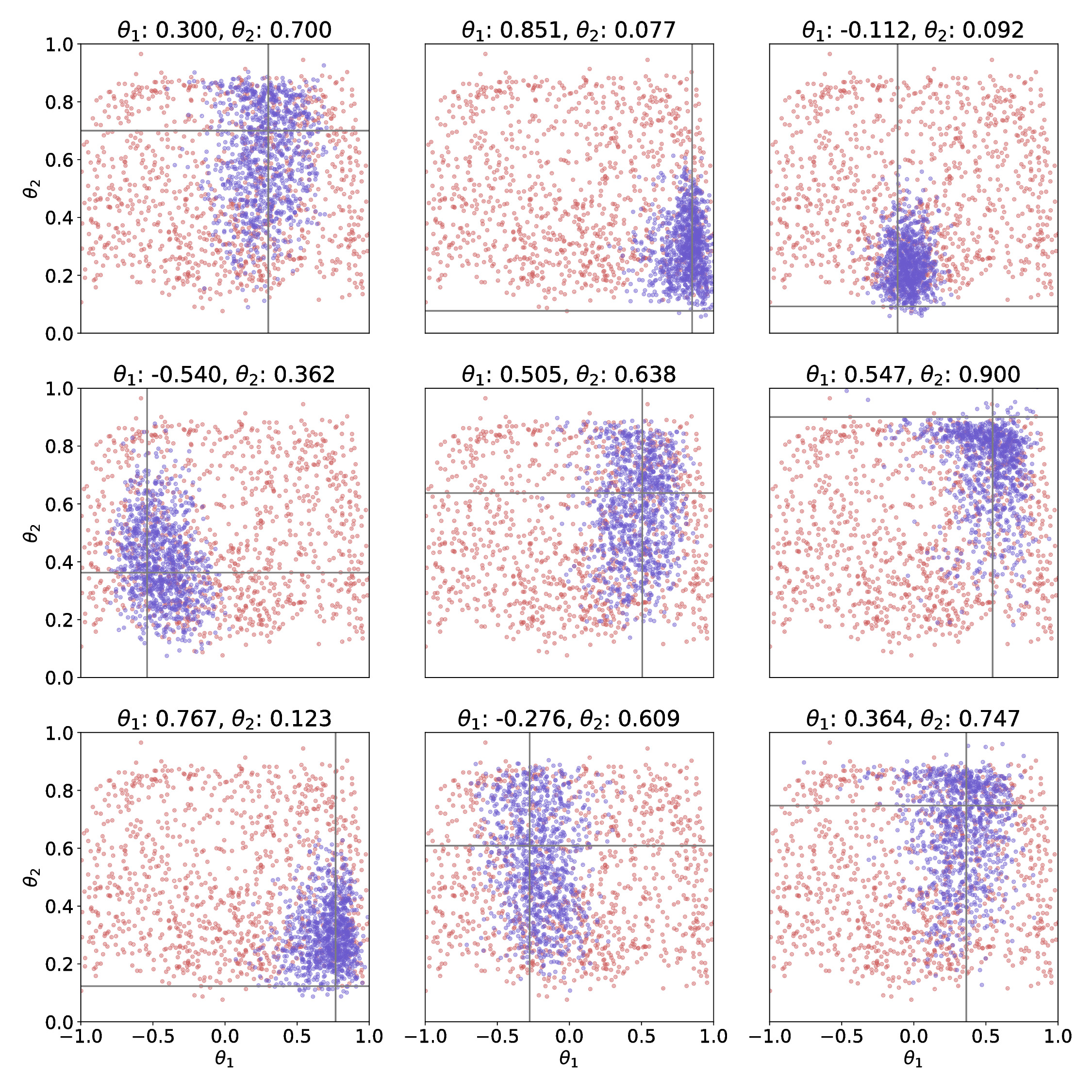}

        \caption{ARCH model: predictions $\bthat(\bx)$ by the deep network of \textcite{jiang_learning_2018}}
        \label{fig:arch_scatterplot_jiang}
    \end{figure}

    Figure \ref{fig:arch_scatterplot_direnet} shows the predictions
    for the DireNet. As in Figure 1 in the main text, the whole
    parameter space is well covered by the predictions for
    $\bthat(\bx)$ sampled from the marginal $p(\bx)$ (in red) while
    the predictions for $\bx$ sampled from $p(\bx \mybar \bt)$ are
    clustered around the particular value of $\bt$ (in blue). Figure
    \ref{fig:arch_scatterplot_jiang} shows the predictions for the
    deep network of \textcite{jiang_learning_2018} in the same
    way. Comparing the two figures, we see that in Figure
    \ref{fig:arch_scatterplot_jiang}, the red points are less spread
    out and the blue points less concentrated than for the DireNet in
    Figure \ref{fig:arch_scatterplot_direnet}. Intuition suggests that
    the data clouds generated by the DireNet are more easily
    classifiable and hence the corresponding LFIRE posteriors also
    more accurate. This is intuition is supported by the
    Kullback-Leibler divergences reported in the main text and the
    additional ones reported in \autoref{tab:app_a_arch_direnet_jiang}.

    Figure \ref{fig:arch_multiple_pred_deep} shows the predictions
    with the additional deep network that we considered. We can here
    see that the predictions are poorer than for the two other neural
    networks. This is possibly due to the reduced flexibility (fewer
    neurons) of the network. The predictions are generally more
    concentrated (less spread-out), both for $\bx \sim p(\bx)$ and
    $\bx \sim p(\bx \mybar \bt)$, than for the proposed DireNet. Table
    \ref{tab:app_a_arch_direnet_jiang} shows that the
    corresponding posteriors are typically less accurate than those
    for the DireNet in line with our intuition. However, it 
    also show that, while less accurate than for the proposed DireNet, the
    posteriors are typically more accurate than those for the deep
    network by \textcite{jiang_learning_2018}. This is possibly due to
    all predictions being compressed into a smaller volume of the
    parameter space, and the subsequent training of the classifier in
    LFIRE being able to accommodate this systematic distortion as pointed out in
    the main text.
        
    Example posteriors are shown in Figures
    \ref{fig:arch_multiple_pred_direnet} to
    \ref{fig:arch_multiple_pred_deep}. The true posterior, calculated
    numerically, is plotted in black for comparison. The figures show
    that the posteriors for the DireNet summary statistics match the
    true one typically better than the posteriors based on the other
    neural network summary statistics.

    \begin{figure}
        \centering
        \includegraphics[width=\textwidth]{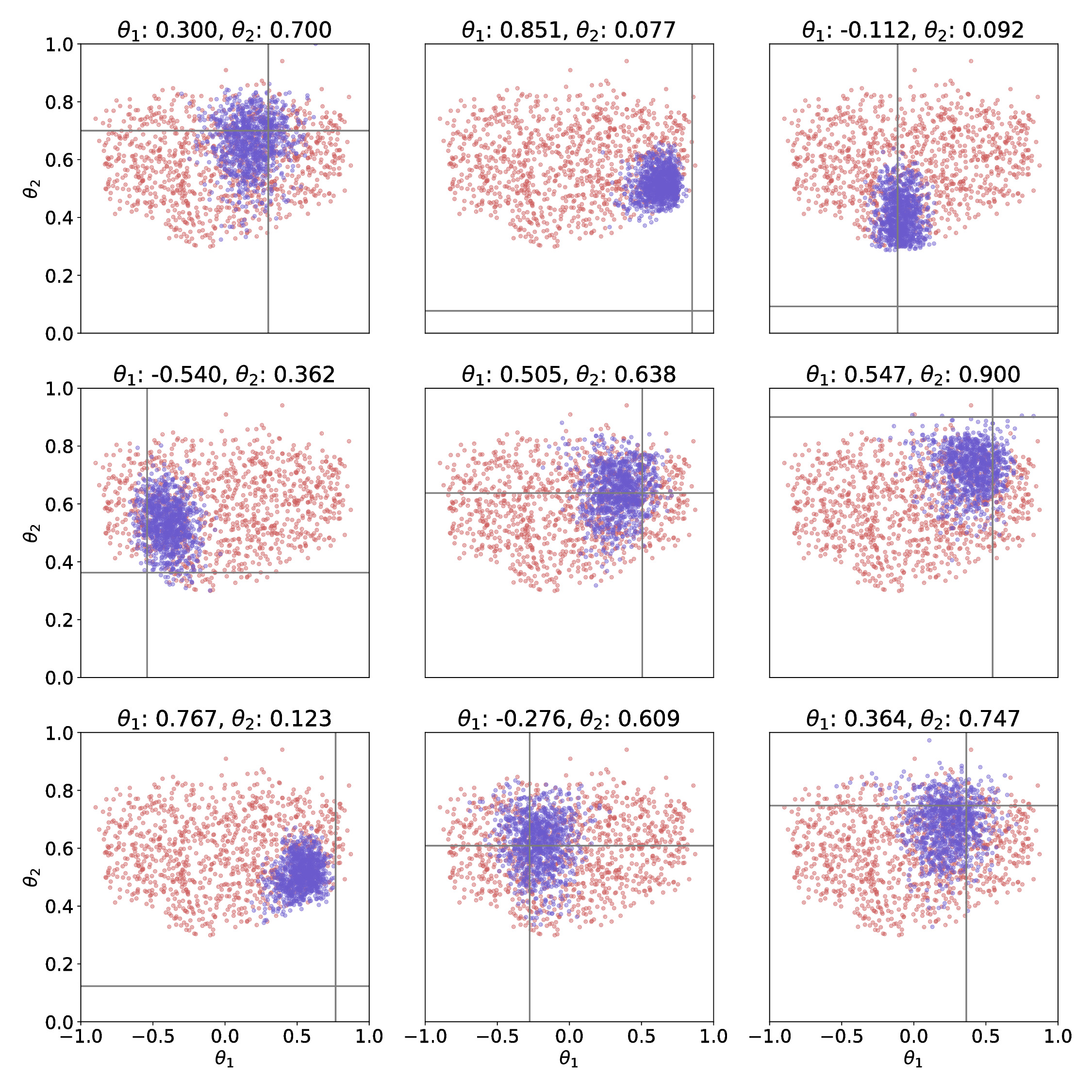}

        \caption{ARCH model: predictions $\bthat(\bx)$ by the additional deep network}
        \label{fig:arch_scatterplot_deep}
    \end{figure}

    \begin{table}
      \centering
      \begin{tabular}{llll}
        \toprule
        Theta &         DireNet &         Alt.\ Deep Net &        \textcite{jiang_learning_2018} \\
        \midrule
        -1.000, 0.250 &  \textbf{0.393} $\pm$ 0.049 &  0.830 $\pm$ 0.105 &  0.840 $\pm$ 0.082 \\
        -1.000, 0.500 &  \textbf{0.503} $\pm$ 0.088 &  1.047 $\pm$ 0.112 &  1.059 $\pm$ 0.086 \\
        -1.000, 0.750 &  \textbf{1.018} $\pm$ 0.162 &  1.135 $\pm$ 0.108 &  1.203 $\pm$ 0.073 \\
        -0.333, 0.000 &  \textbf{0.287} $\pm$ 0.050 &  0.975 $\pm$ 0.120 &  0.951 $\pm$ 0.094 \\
        -0.333, 0.250 &  \textbf{0.343} $\pm$ 0.059 &  0.743 $\pm$ 0.128 &  0.887 $\pm$ 0.129 \\
        -0.333, 0.500 &  \textbf{0.673} $\pm$ 0.114 &  0.824 $\pm$ 0.086 &  0.889 $\pm$ 0.078 \\
        -0.333, 0.750 &  \textbf{0.508} $\pm$ 0.073 &  0.971 $\pm$ 0.164 &  1.006 $\pm$ 0.093 \\
        -0.333, 1.000 &  \textbf{0.751} $\pm$ 0.093 &  1.111 $\pm$ 0.125 &  1.102 $\pm$ 0.084 \\
        0.333, 0.000 &  \textbf{0.373} $\pm$ 0.113 &  0.705 $\pm$ 0.125 &  0.847 $\pm$ 0.115 \\
        0.333, 0.250 &  \textbf{0.411} $\pm$ 0.062 &  0.749 $\pm$ 0.094 &  0.753 $\pm$ 0.102 \\
        0.333, 0.500 &  \textbf{0.393} $\pm$ 0.042 &  0.869 $\pm$ 0.133 &  0.809 $\pm$ 0.116 \\
        0.333, 0.750 &  \textbf{0.695} $\pm$ 0.095 &  1.062 $\pm$ 0.103 &  0.943 $\pm$ 0.078 \\
        0.333, 1.000 &  \textbf{0.738} $\pm$ 0.094 &  1.013 $\pm$ 0.082 &  1.096 $\pm$ 0.084 \\
        1.000, 0.250 &  \textbf{0.698} $\pm$ 0.065 &  0.839 $\pm$ 0.094 &  1.185 $\pm$ 0.130 \\
        1.000, 0.500 &  \textbf{0.835} $\pm$ 0.041 &  0.970 $\pm$ 0.061 &  1.257 $\pm$ 0.064 \\
        1.000, 0.750 &  \textbf{1.044} $\pm$ 0.089 &  1.372 $\pm$ 0.120 &  1.452 $\pm$ 0.114 \\
        \bottomrule
      \end{tabular}
      \caption[ARCH model; all results]{ARCH model. Average KL divergence and standard errors over 20 runs for data sets generated with different values of $\bt$. The DireNet performs better than both other approaches. The KL divergence is computed between the true posterior and the estimated one.}
      \label{tab:app_a_arch_direnet_jiang}
    \end{table}
    
    \begin{figure}
      \centering
      \includegraphics[width=\textwidth]{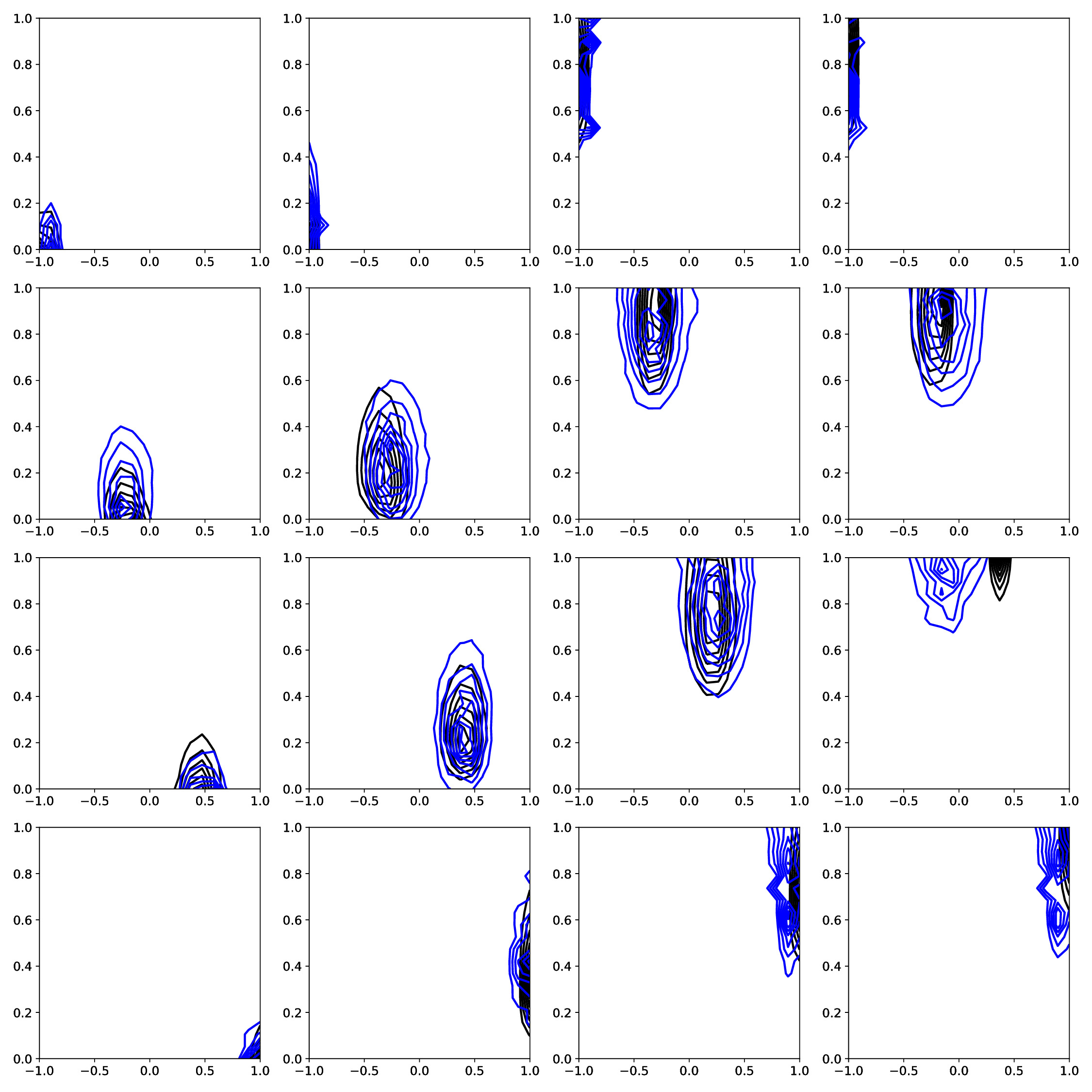}
      
        \caption{ARCH model: posteriors computed using the DireNet summary statistics.}
        \label{fig:arch_multiple_pred_direnet}
    \end{figure}

    \begin{figure}
      \centering
      \includegraphics[width=\textwidth]{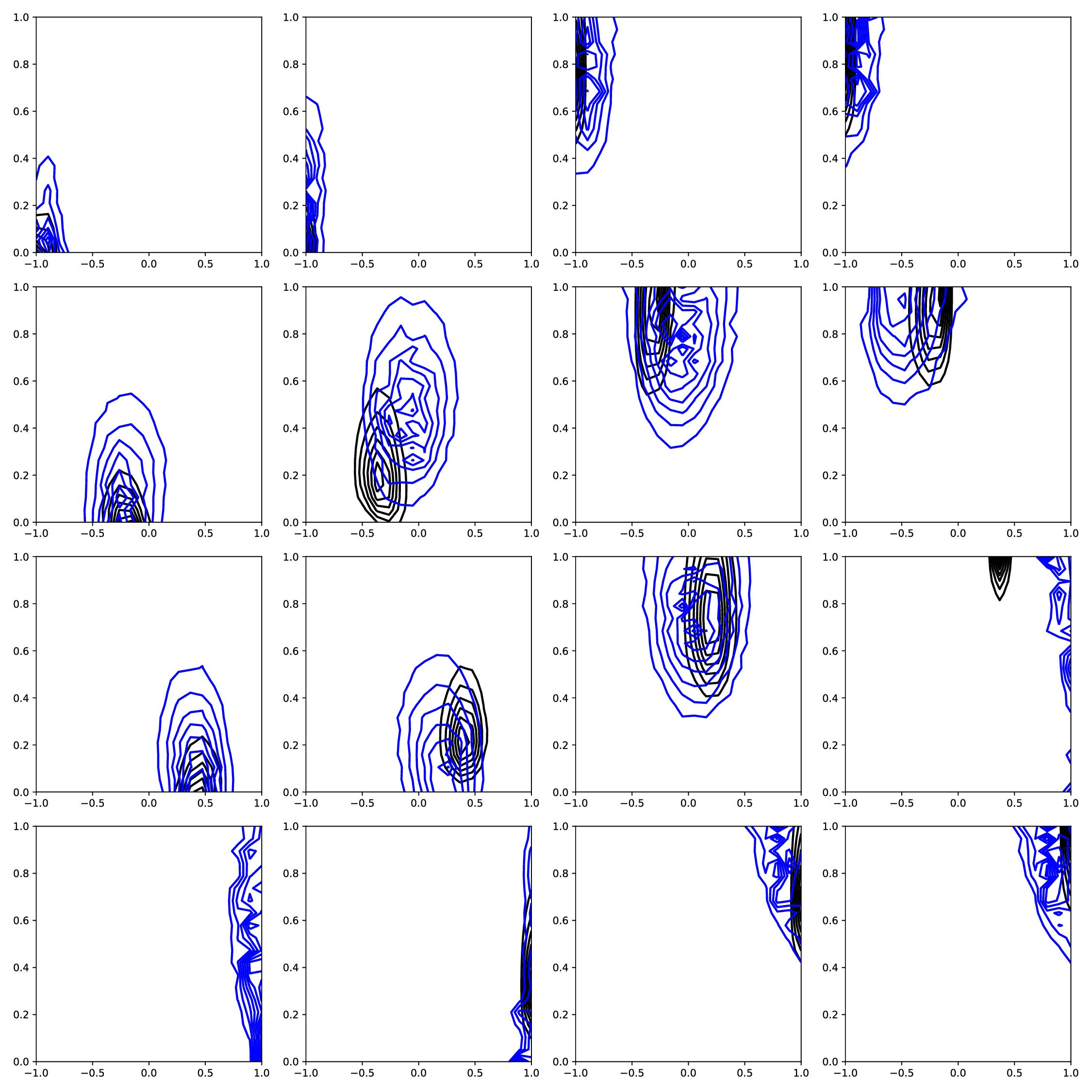}
      
      \caption{ARCH model: posteriors computed with summary statistics given by the deep network of \textcite{jiang_learning_2018}.}
      \label{fig:arch_multiple_pred_jiang}
    \end{figure}
    
    \begin{figure}
      \centering
      \includegraphics[width=\textwidth]{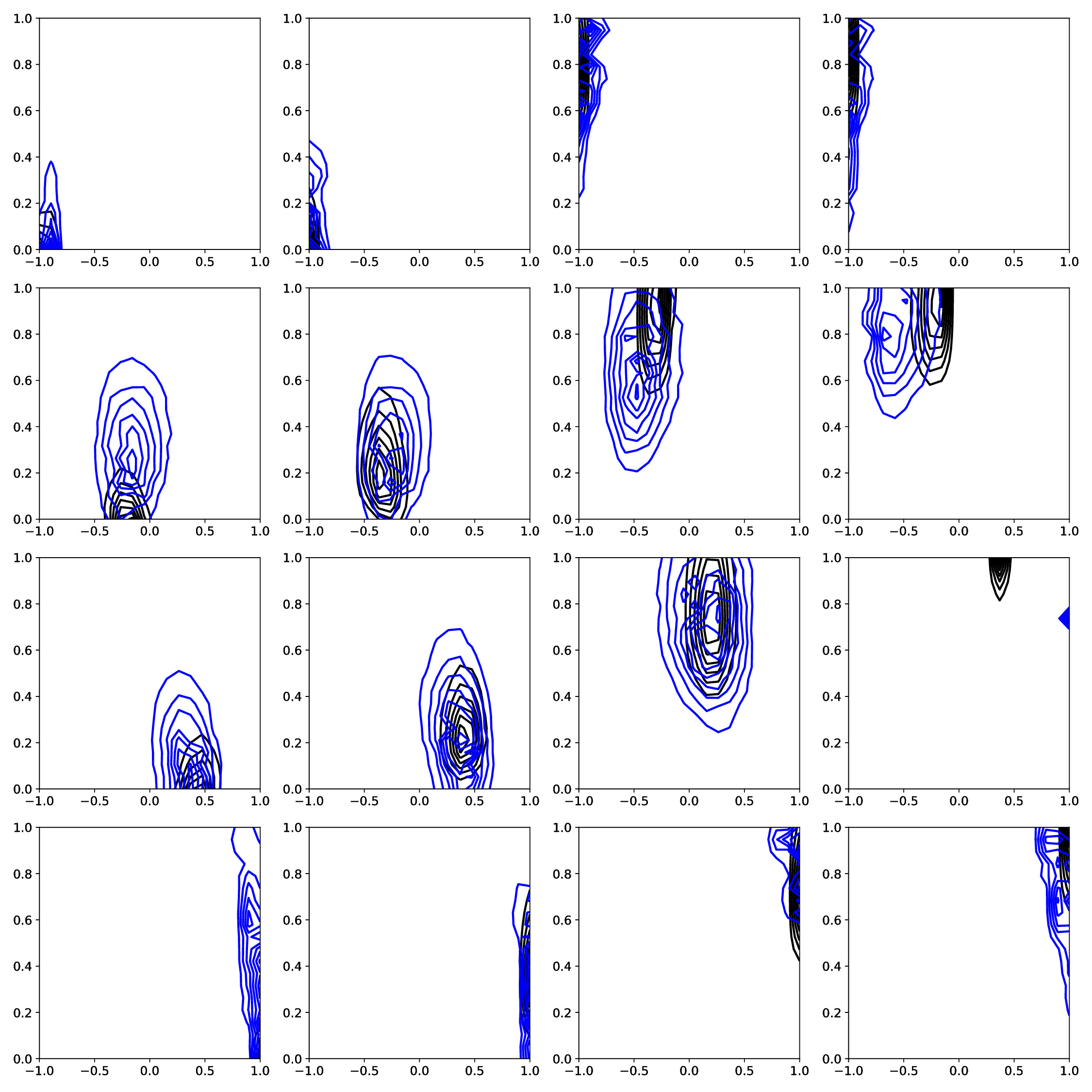}
      
      \caption{ARCH model: posteriors computed with summary statistics given by the additional deep network.}
      \label{fig:arch_multiple_pred_deep}
    \end{figure}

    \clearpage

    \subsection{Lotka-Volterra model}

The lower and upper limit of the 95\% confidence interval for the mean
of $\Delta^\text{rel}_i$ computed via bootstrapping, as well as the
average of the bootstrap distribution of the mean is shown in
\autoref{tab:lv_bootstrap}. The results are based on 200 bootstrapped
data sets (of size 500).
The bootstrap confidence
intervals are on the negative axis which indicates better performance
of the proposed method (see main text).

\begin{figure}[h!]
        \centering
        \includegraphics[width=0.9\textwidth]{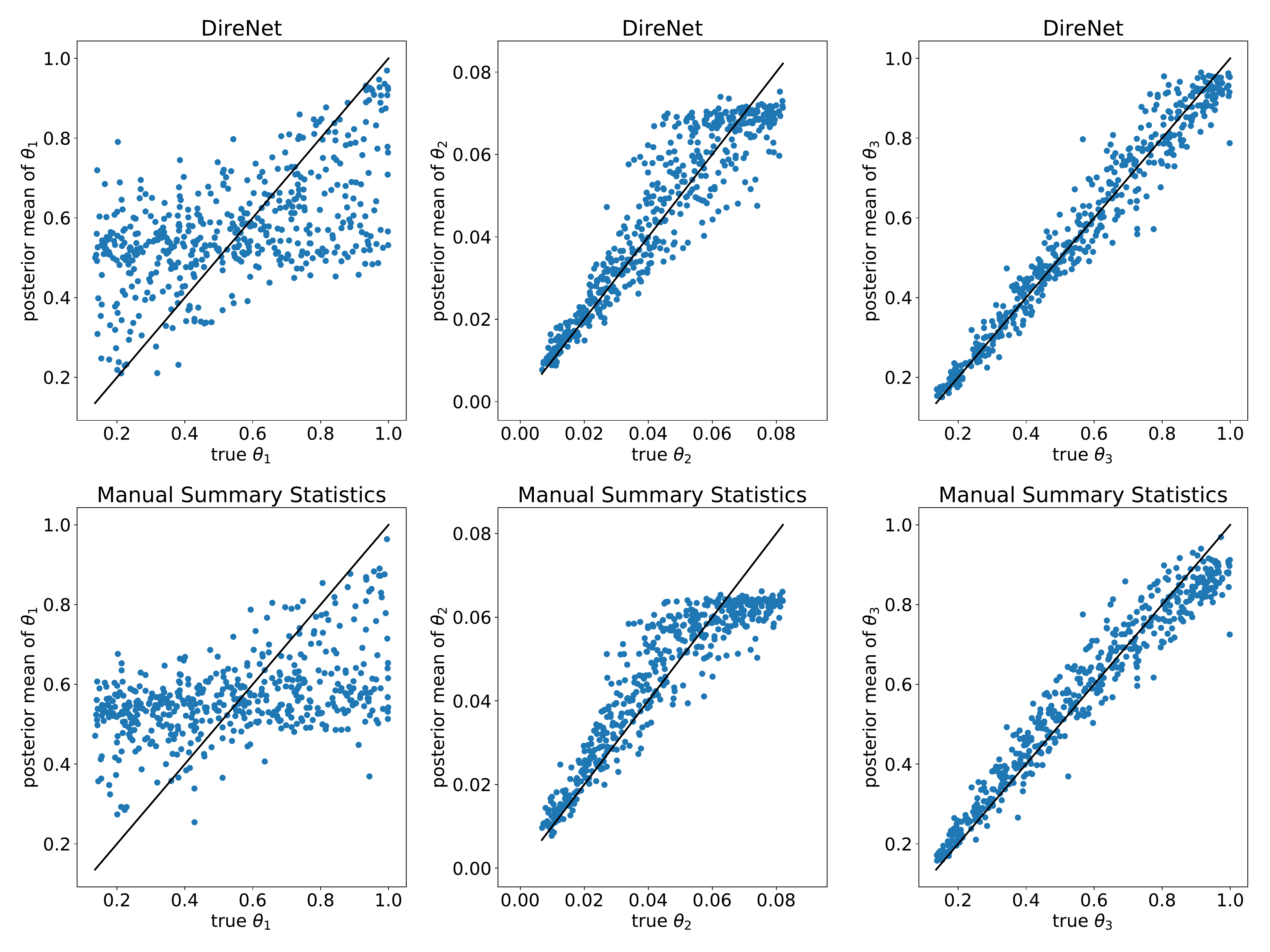}        
        \caption{Lotka-Volterra model: true parameter values vs. posterior means comparing DireNet to manual summary statistics.}
        \label{fig:trupred_lv}
    \end{figure}

\autoref{fig:trupred_lv} plots true parameter values versus the
posterior means for the $500$ inference problems, when using the
DireNet and the manual summary statistics in LFIRE. We see that the
posterior means when using manual summary statistics are more over- or
underestimated as compared to the case of the DireNet.

 \begin{table}[h!]
        \centering
        \begin{tabular}{llll}
            \toprule
                     Parameter & Lower Limit & Upper Limit & Average \\
            \midrule
                $\theta_1$ & -0.052 & -0.026 & -0.039 \\
                $\theta_2$ & -0.039 & -0.020 & -0.029 \\
                $\theta_3$ & -0.034 & -0.023 & -0.028 \\
            \bottomrule
            \end{tabular}
        \caption{Lotka-Volterra model. Bootstrap distribution of the
          mean of $\Delta^\text{rel}_i$, showing the lower and upper
          limit of the 95\% intervals as well as the average value.}
        \label{tab:lv_bootstrap}
    \end{table}

    \clearpage
    \subsection{Ricker model} \vspace{-0.5ex}
We show in Tables \ref{tab:ricker_wood_bootstrap} and
\ref{tab:ricker_wood_bootstrap_med} the bootstrap results for the
comparison between LFIRE with the DireNet summary statistics and the
synthetic likelihood with original summary statistics by
\citet{Wood2010}. For $\sigma$, the bootstrap confidence intervals for
the mean of $\Delta^\text{rel}_i$ is on the positive axis while for the
median on the negative axis (see main text). Tables
\ref{tab:ricker_semiautoabc_bootstrap} and
\ref{tab:ricker_semiautoabc_bootstrap_med} shows the corresponding
bootstrap results for the comparison to semi-automatic ABC
\citep{Fearnhead2012}. Here, the confidence interval for the
median of $\Delta^\text{rel}_i$ for $\log r$ just includes zero while the corresponding interval for the mean is negative, which
indicates a small difference in performance.

\begin{figure}
        \centering
        \includegraphics[width=\textwidth]{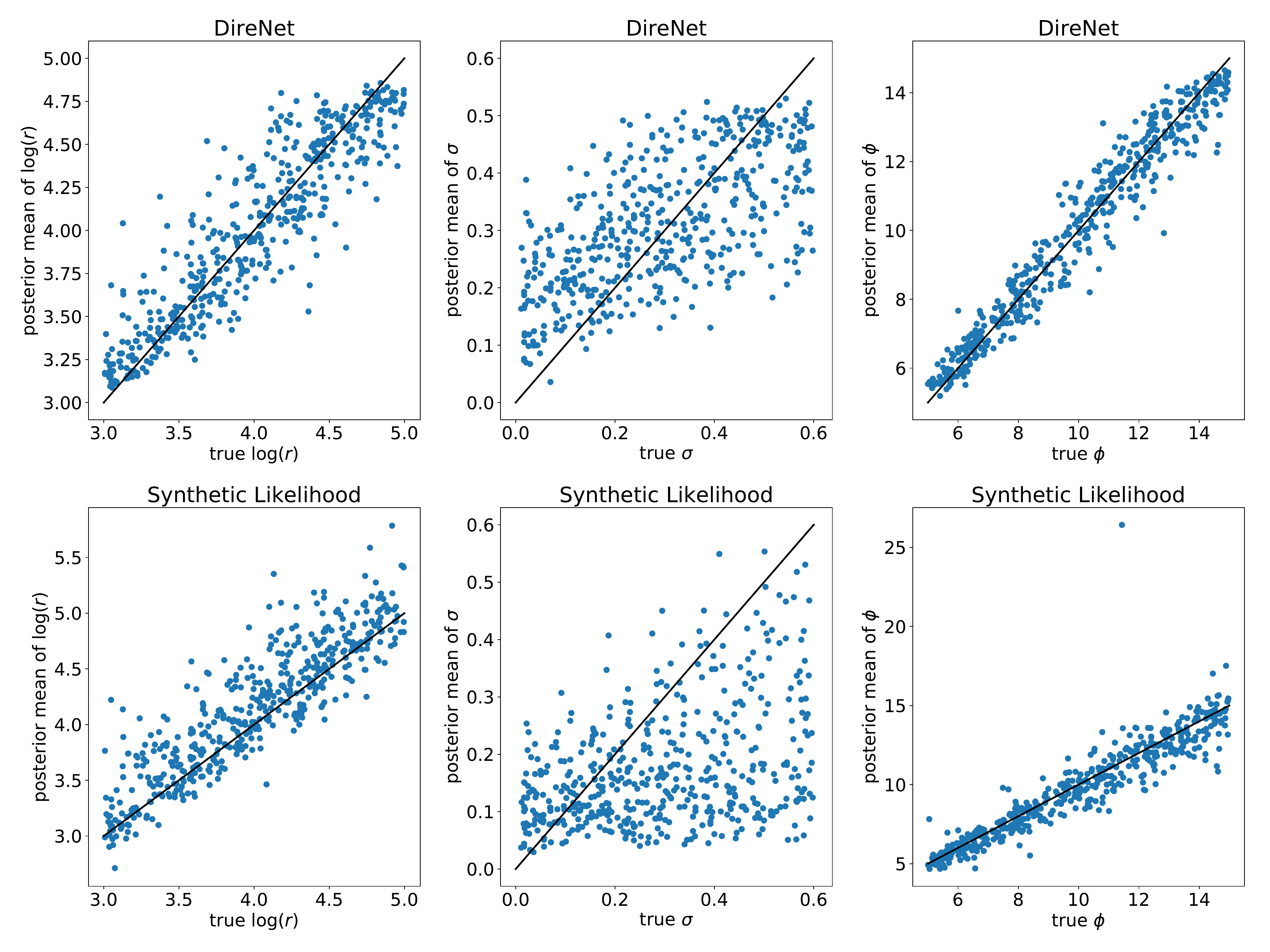}

        \caption{Ricker model: true parameter values vs. posterior
          means comparing LFIRE with DireNet to synthetic likelihood.}
        \label{fig:trupred_ricker_synlik}
    \end{figure}

\begin{figure}
        \centering
        \includegraphics[width=\textwidth]{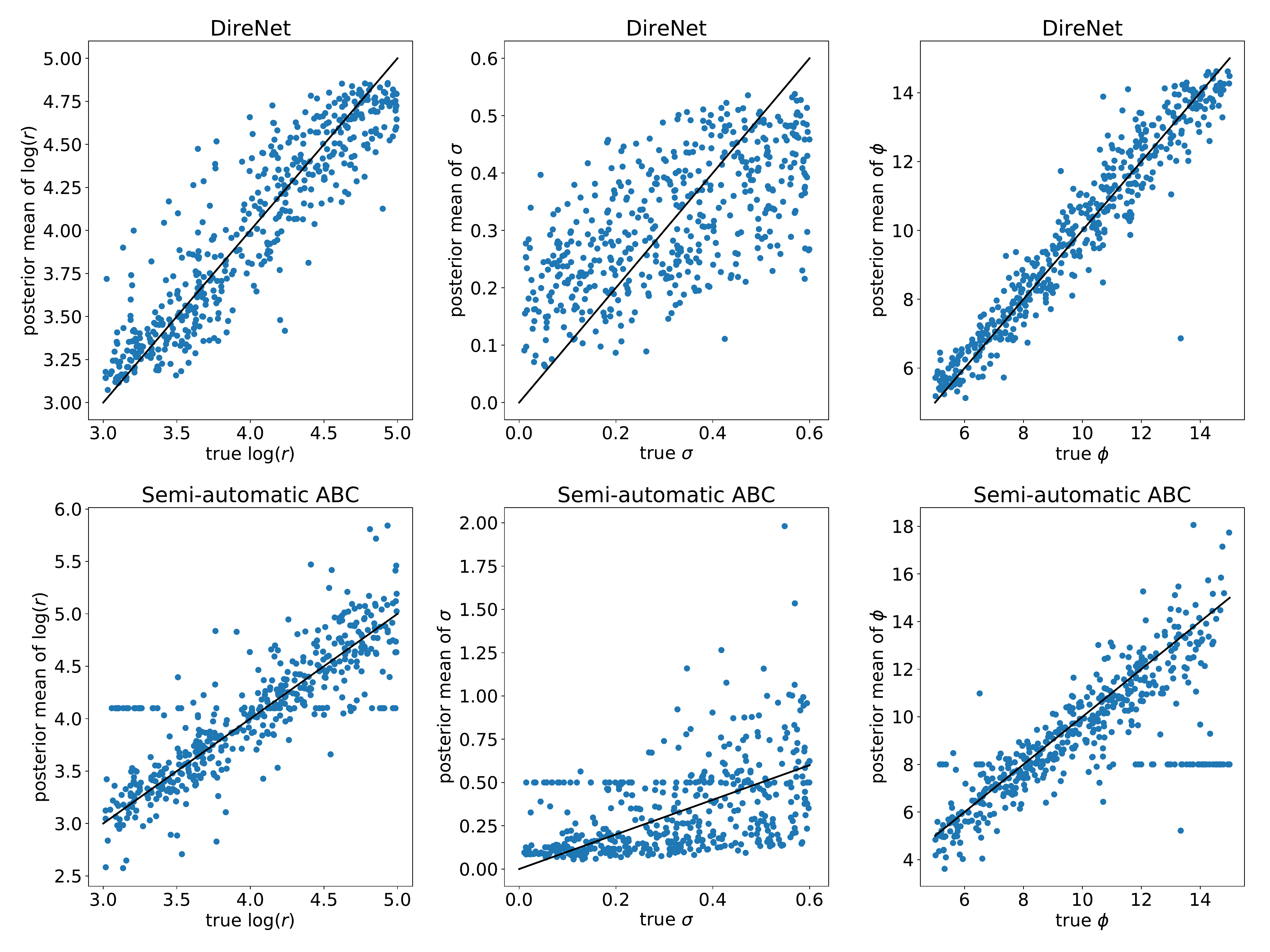}

        \caption{Ricker model: true parameter values vs. posterior
          means comparing LFIRE with DireNet to semi-automatic ABC.}
        \label{fig:trupred_ricker_semiauto}
    \end{figure}

\autoref{fig:trupred_ricker_synlik} and
\autoref{fig:trupred_ricker_semiauto} plot the true parameters versus
the posterior means for the different methods. In general, the scatter plots for LFIRE with the DireNet are more concentrated around the diagonal than the alternative methods, indicating better overall performance.

    \begin{table}[h!]
      \centering
      \begin{tabular}{llll}
        \toprule
        Parameter & Lower Limit & Upper Limit & Average \\
          \midrule
          $\log{r}$ & -0.018 & -0.010 & -0.014 \\
          $\sigma$ & \ 0.122 &  \ 0.376 &  \ 0.237 \\
          $\phi$ & -0.024 & -0.011 & -0.017 \\
          \bottomrule
      \end{tabular}
      \caption{Ricker model: comparing LFIRE with the DireNet to \emph{synthetic
        likelihood}. Bootstrap results for the \emph{mean} of $\Delta^\text{rel}_i$ computed as in Table
        \ref{tab:lv_bootstrap}.
        \label{tab:ricker_wood_bootstrap}}
    \end{table}
    \vspace{-0.5ex}   
    \begin{table}[h!]
      \centering
      \begin{tabular}{llll}
        \toprule
        Parameter & Lower Limit & Upper Limit & Average \\
        \midrule
        $\log{r}$ & -0.011 & -0.006 & -0.008 \\
        $\sigma$ & -0.165 & -0.103 & -0.135 \\
        $\phi$ & -0.011 & -0.002 & -0.007 \\
        \bottomrule
      \end{tabular}
      \caption{Ricker model: comparing LFIRE with the DireNet to
        \emph{synthetic likelihood}. Bootstrap distribution of the \emph{median}
          of $\Delta^\text{rel}_i$, showing the lower and upper limit
          of the 95\% intervals as well as the average
          value.\label{tab:ricker_wood_bootstrap_med}}
    \end{table}
    \vspace{-0.5ex}   
    \begin{table}[h!]
      \centering
      \begin{tabular}{llll}
        \toprule
        Parameter & Lower Limit & Upper Limit & Average\\
        \midrule
        $\log{r}$ & -0.022 & -0.009 & -0.016 \\
        $\sigma$ & -0.290 & \ 0.093 & -0.082 \\
        $\phi$ & -0.067 & -0.043 & -0.055 \\
        \bottomrule
      \end{tabular}
        \caption{Ricker model: comparing LFIRE with the DireNet to
          \emph{semi-automatic ABC}. Bootstrap results for the \emph{mean} of
          $\Delta^\text{rel}_i$ computed as in Table
          \ref{tab:lv_bootstrap}.\label{tab:ricker_semiautoabc_bootstrap}}
        \vspace{0.1ex}   
      \centering
      \begin{tabular}{llll}
        \toprule
        Parameter & Lower Limit & Upper Limit & Average \\
        \midrule
        $\log{r}$ & -0.005 &  0.003 & -0.001 \\
        $\sigma$ & -0.193 & -0.101 & -0.147 \\
        $\phi$ & -0.030 & -0.013 & -0.023 \\
        \bottomrule
      \end{tabular}
      \caption{Ricker model: comparing LFIRE with the DireNet to
        \emph{semi-automatic ABC}. Bootstrap results for the \emph{median} of
        $\Delta^\text{rel}_i$ computed as in Table
        \ref{tab:ricker_wood_bootstrap_med}. \label{tab:ricker_semiautoabc_bootstrap_med}} 
    \end{table}

   \clearpage
    
    \subsection{Lorenz model}

\begin{figure}[h!]
        \centering
        \includegraphics[width=0.8\textwidth]{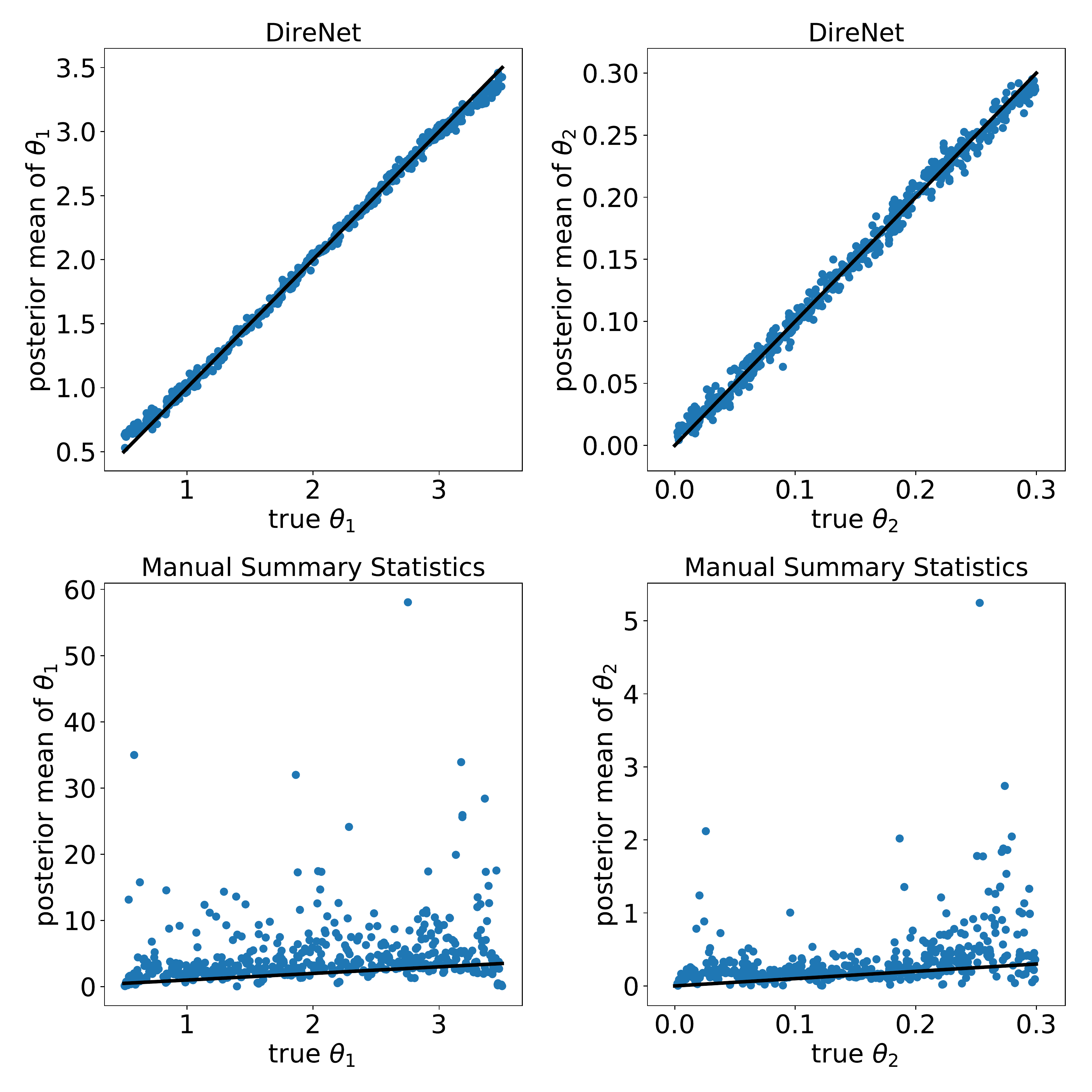}

        \caption{Lorenz model: true parameter values vs. posterior
          means comparing DireNet with manual summary statistics.}
        \label{fig:trupred_lorenz}
    \end{figure}

\begin{figure}[t!]
      \centering
      \begin{subfigure}[t]{0.48\textwidth}
        \centering
        \includegraphics[width=\textwidth]{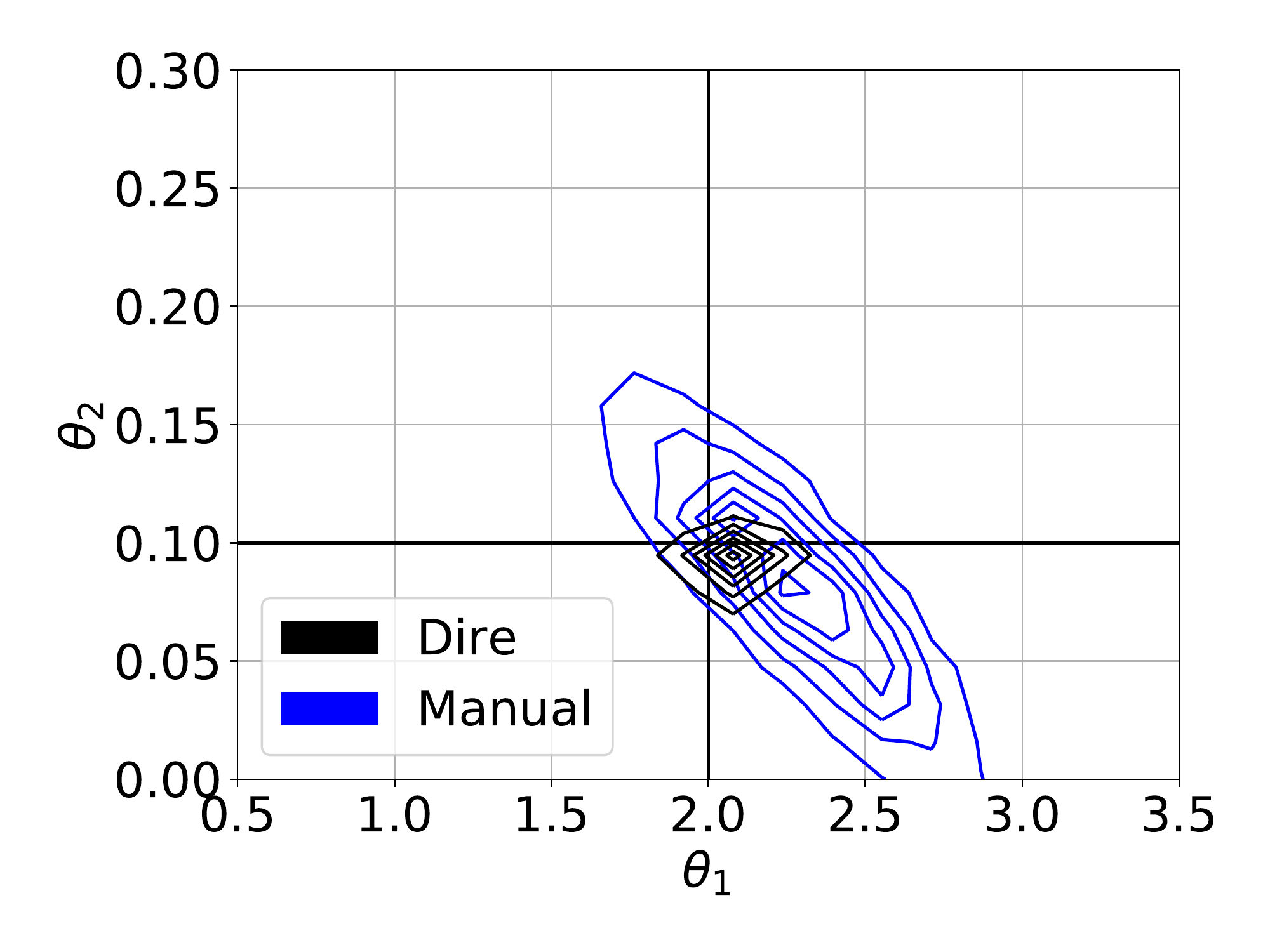}
        \caption{Obs data generated with $\bt=(2.0, 0.1)$}
      \end{subfigure}
      \begin{subfigure}[t]{0.48\textwidth}
        \centering
        \includegraphics[width=\textwidth]{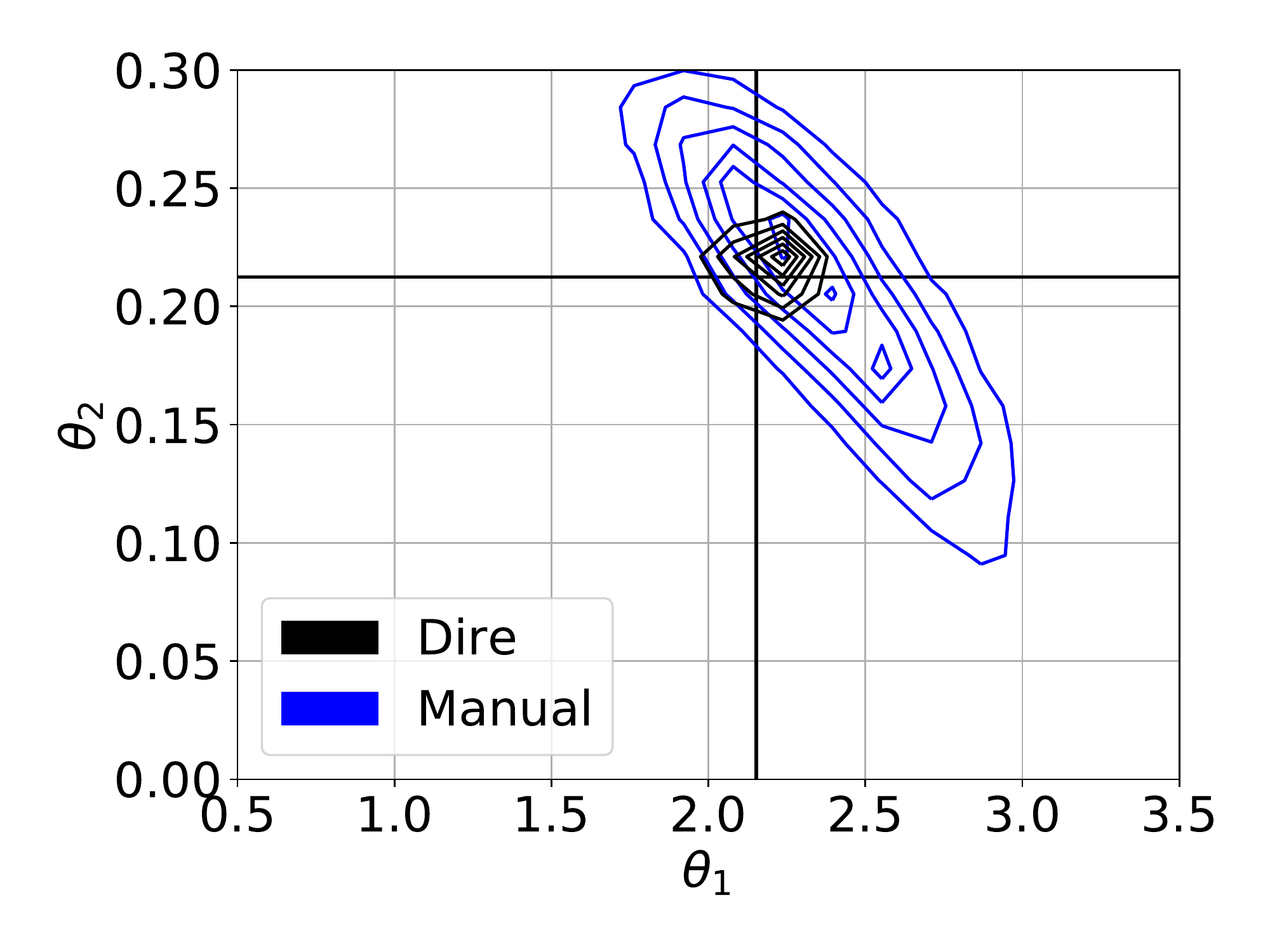}
        \caption{Obs data generated with $\bt = (2.152, 0.212)$}
      \end{subfigure}
        \caption{Lorenz model: posteriors for data sets generated with
          two different parameter values. \label{fig:lorenz_posterior}}
    \end{figure}

   \autoref{tab:lorenz_bootstrap} shows the bootstrap results for the
   mean of $\Delta^\text{rel}_i$, and Figure \ref{fig:trupred_lorenz} shows
   scatter plots of the true parameter value versus the posterior
   means. As mentioned in the main text, we see here a large
   improvement over the manual summary statistics. Given the large
   improvement, we also show in \autoref{fig:lorenz_posterior} example
   posterior distributions. Previous work \citep{Dutta2016} showed the
   posterior for $\bt=(2.0, 0.1)$ and our posterior with manual
   summary statistics matches their result well. Importantly, we see
   that when using DireNet, the posteriors are much tighter and
   centered close to true data generating parameter.
    
    \begin{table}[h!]
        \centering
        \begin{tabular}{llll}
            \toprule
                     Parameter & Lower Limit & Upper Limit & Average \\
            \midrule
                $\theta_1$ & -2.116 & -1.465 & -1.725 \\
                $\theta_2$ & -3.194 & -2.126 & -2.589 \\
            \bottomrule
            \end{tabular}
        \caption{Lorenz model: comparing DireNet to manual summary
          statistics. Bootstrap distribution for the mean
          $\Delta^\text{rel}_i$ computed as in Table
          \ref{tab:lv_bootstrap}.}
        \label{tab:lorenz_bootstrap}
    \end{table}

 